%% file: acl_latex.tex
\pdfoutput=1

\documentclass[11pt]{article}

\usepackage[preprint]{acl}

\usepackage{times}
\usepackage{latexsym}

\usepackage[T1]{fontenc}

\usepackage[utf8]{inputenc}

\usepackage{microtype}

\usepackage{inconsolata}

\usepackage{graphicx}
\usepackage{microtype}
\usepackage{graphicx}
\usepackage{subfigure}
\usepackage{booktabs} 

\usepackage{hyperref}
\usepackage{tabularx}

\usepackage{amsmath}
\usepackage{amssymb}
\usepackage{mathtools}
\usepackage{amsthm}

\usepackage[capitalize,noabbrev]{cleveref}

\theoremstyle{plain}

\theoremstyle{definition}

\theoremstyle{remark}

\usepackage[textsize=tiny]{todonotes}

\usepackage[utf8]{inputenc} 
\usepackage[T1]{fontenc}    
\usepackage{hyperref}       
\usepackage{url}            
\usepackage{booktabs}       
\usepackage{amsfonts}       
\usepackage{nicefrac}       
\usepackage{microtype}      
\usepackage{xcolor}         
\definecolor{darkgreen}{rgb}{0,0.6,0} 
\usepackage{longtable}
\usepackage{graphicx}
\usepackage{array}
\usepackage{multirow}
\usepackage{xcolor,colortbl}
\usepackage{xspace}
\usepackage{makecell}
\usepackage{boldline}
\usepackage{pifont}
\usepackage{tabularx}
\usepackage{graphicx}
\usepackage{wrapfig}
\usepackage[export]{adjustbox}
\usepackage{fancyvrb}
\usepackage{fvextra}
\usepackage[most,breakable]{tcolorbox}
\usepackage{caption}
\usepackage{pgfplots}
\pgfplotsset{compat=newest}
\usepgfplotslibrary{units}
\usepackage{placeins}
\usepackage{lineno}
\usepackage{enumitem}

\usepackage{listings}

\definecolor{codegreen}{rgb}{0,0.6,0}
\definecolor{codegray}{rgb}{0.5,0.5,0.5}
\definecolor{codepurple}{rgb}{0.58,0,0.82}

\lstdefinestyle{pddl}{
    language=Lisp, 
    basicstyle=\ttfamily\footnotesize,
    commentstyle=\color{codegreen},
    keywordstyle=\color{magenta},
    numberstyle=\tiny\color{codegray},
    stringstyle=\color{codepurple},
    breakatwhitespace=false,         
    breaklines=true,                 
    captionpos=b,                    
    keepspaces=true,                 
    numbersep=5pt,                  
    showspaces=false,                
    showstringspaces=false,
    showtabs=false,                  
    tabsize=2,
    morekeywords={define, domain, requirements, predicates, parameters, precondition, effect, action, :, and, not},
    deletekeywords={}, 
    morecomment=[l]{;} 
}

\newcommand{\benchmark}[0]{\textsc{Text2World}\xspace}
\newcommand{\brieftitle}[0]{\textsc{Text2World}\xspace}
\newcommand{\exec}[0]{\textsc{Exec.}\xspace}
\newcommand{\simmetric}[0]{\textsc{Sim.}\xspace}
\newcommand{\fonepred}[0]{\textsc{F1\textsubscript{pred}}\xspace}
\newcommand{\foneparam}[0]{\textsc{F1\textsubscript{param}}\xspace}
\newcommand{\foneprecond}[0]{\textsc{F1\textsubscript{precond}}\xspace}
\newcommand{\foneeff}[0]{\textsc{F1\textsubscript{eff}}\xspace}

\begin{document}


\title{%
  \raisebox{-0.35ex}{\includegraphics[width=0.55cm]{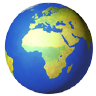}}%
  \hspace{3.5px}\brieftitle: Benchmarking Large Language Models for \\ Symbolic World Model Generation%
}



\author{Mengkang Hu\footnotemark[1]$^{\spadesuit}$ \quad Tianxing Chen\footnotemark[1]$^{\spadesuit}$ \quad Yude Zou\thanks{~Equal Contribution. $\dagger$~Corresponding Author}$^{\clubsuit}$ \quad Yuheng Lei$^{\spadesuit}$ \quad Qiguang Chen$^{\heartsuit}$ \\
	\textbf{Ming Li$^{\spadesuit}$ \quad Yao Mu$^\spadesuit$ \quad Hongyuan Zhang$^\spadesuit$ \quad Wenqi Shao$^\diamondsuit$ \quad Ping Luo$^\dagger$$^\spadesuit$} \\
	$^{\spadesuit}$ The University of Hong Kong \quad $^{\clubsuit}$  Shenzhen University \\
        $^{\heartsuit}$ Harbin Institute of Technology \quad
	$^{\diamondsuit}$ Shanghai AI Laboratory\\
	\texttt{mkhu@connect.hku.hk}, \quad \texttt{pluo.lhi@gmail.com}\\
    \href{https://text-to-world.github.io/}{text-to-world.github.io}  
    }


\maketitle
\begin{abstract}
Recently, there has been growing interest in leveraging large language models (LLMs) to generate symbolic world models from textual descriptions. 
Although LLMs have been extensively explored in the context of world modeling, prior studies encountered several challenges, including evaluation randomness, dependence on indirect metrics, and a limited domain scope.
To address these limitations, we introduce a novel benchmark, \benchmark, based on planning domain definition language (PDDL), featuring hundreds of diverse domains and employing multi-criteria, execution-based metrics for a more robust evaluation.
We benchmark current LLMs using \benchmark and find that reasoning models trained with large-scale reinforcement learning outperform others.
However, even the best-performing model still demonstrates limited capabilities in world modeling.
Building on these insights, we examine several promising strategies to enhance the world modeling capabilities of LLMs, including test-time scaling, agent training, and more.
We hope that \benchmark can serve as a crucial resource, laying the groundwork for future research in leveraging LLMs as world models.
\end{abstract}

\section{Introduction}
\label{sec:intro}

The significance of world models for intelligent behavior has been historically acknowledged in early psychological theories, which posited that organisms employ internal representations of the external world for prediction and planning~\citep{craik1967nature}.
Furthermore, ~\citet{lecun2022path} extends this concept by highlighting world modeling as a core component of autonomous machine intelligence.
In this paper, we primarily study \textit{symbolic world models} (also known as domain models), which are formal representations of an environment’s dynamics and constraints.
In recent years, Large Language Models (LLMs)~\cite{opeiai2022gpt,yang2024qwen2,meta2024llama3} have showcased their understanding of common-world knowledge, making them promising candidates for generating symbolic world models, which requires inferring action dynamics and constraints from solely natural language description.
Some works have already explored this across numerous tasks, including planning~\citep{hu2024agentgen, guan2023leveraging}, game design~\citep{wang2023bytesized32,wang2024can}, reinforcement learning~\cite{tang2024worldcoder} among others.

Despite extensive exploration, previous work for evaluating symbolic world model generation suffers from several key limitations:
\textit{(i)} \textit{\textbf{Limited Domain Scope}}: These studies are often confined to a narrow set of domains (typically fewer than 20), which limits the generalizability and applicability of their findings~\cite{oswald2024large,silver2024generalized,wong2023word}.
\textit{(ii)} \textit{\textbf{Evaluation Randomness}}: Some works rely on LLM-based evaluation methods, which may introduce additional margins of error~\cite{wang2023bytesized32}. 
Preliminary experiments in Section~\ref{sec:llmself} demonstrate that the LLM-based evaluation exhibits a low inter-annotator agreement with human annotators (Cohen’s $\kappa=0.10$).
\textit{(iii)} \textit{\textbf{Indirect Evaluation}}: Some studies evaluate world models based on end-to-end success rates in model-based planning, making it difficult to identify specific failure modes~\cite{guan2023leveraging,dainese2024generating}.

\begin{figure*}[htbp]
    \centering
    \includegraphics[width=\linewidth]{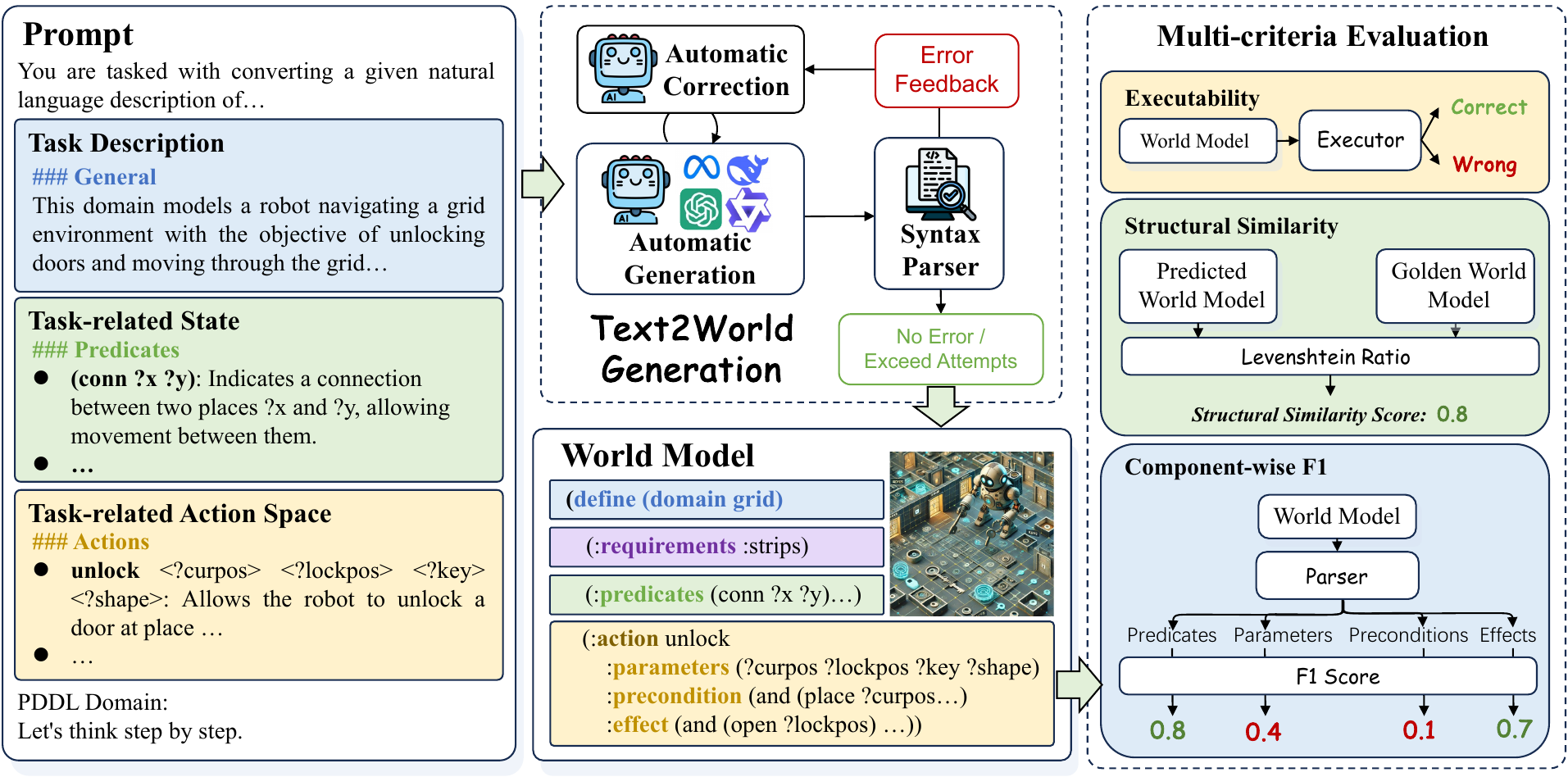}
    \caption{
    Overview of \benchmark.
    }
    \label{fig:main}
\end{figure*}

Motivated by these issues, this paper introduces a novel benchmark \benchmark based on the Planning Domain Definition Language (PDDL) as illustrated in Figure~\ref{fig:main}.
Specifically, to address the first issue, we initially gathered a broad set of domains, which were then filtered through an automated pipeline and manually curated to ensure their quality, ultimately resulting in a collection of hundreds of diverse domains.
Furthermore, to tackle the second issue, we designed multi-criteria, execution-based metrics to ensure a more robust assessment. Specifically, we not only employed structural similarity for an overall evaluation but also designed component-wise F1 scores to assess finer-grained aspects such as action dynamics.
Moreover, to overcome the third issue, we systematically apply these metrics to assess the generated world model directly, eliminating reliance on indirect feedback mechanisms.

We also performed data contamination analysis using n-gram matching~\cite{touvron2023llama}, revealing a lower contamination rate ($\mu$ = 0.04) compared to prior works~\citep{guan2023leveraging,smirnov2024generating}, indicating that \benchmark effectively evaluates LLMs' world modeling capabilities rather than pattern memorization.


We used \benchmark to benchmark the world modeling capabilities of 16 different LLMs from 9 model families. 
Experimental results in Table~\ref{tab:main} highlight several key findings: 
\textit{(i)} \textit{The most advanced LLMs still struggle with \benchmark;}
\textit{(ii)} \textit{large reasoning models trained by reinforcement learning show stronger world modeling capabilities;}
and \textit{(iii)} \textit{error correction significantly improves model performance.}
To gain a deeper understanding, we performed a manual error analysis and found that the majority were due to the LLMs' inability to include essential preconditions or effects. 
We also explored several strategies to enhance the world modeling capabilities of LLMs. 
Specifically, we initially experimented with scaling the test-time budget and observed consistent improvements as the test-time budget increased.  
Additionally, methods like fine-tuning and in-context learning contributed positively to model effectiveness.
Moreover, we found that supervised fine-tuning on agent trajectory data yielded unexpected gains, underscoring the importance of robust world modeling for developing high-performing agents.

To facilitate further research, benchmark and code are available at  \href{https://text-to-world.github.io/}{this URL}.

\section{Preliminary}


\subsection{World Model}

We formally define a symbolic world model as $D = \langle F, A \rangle$, where $F$ represents the set of fluents (state variables represented as predicates) and $A$ is the set of possible actions. 
Each fluent $f \in F$ is a predicate of the form $p(x_1, ..., x_n)$, where $p$ is the predicate name and $x_1, ..., x_n$ are typed variables. 
Each action $a \in A$ is defined as a tuple $a = \langle \alpha, \mathcal{P}, \varphi, \mathcal{E} \rangle$ where: 
i) $\alpha$ denotes the action signature (identifier); 
ii) $\mathcal{P}$ represents a list of typed parameters $(p_1, ..., p_k)$; 
iii) $\varphi$ specifies the preconditions: a logical formula over fluents that must hold for the action to be applicable; 
and iv) $\mathcal{E}$ defines the effects: a set of fluent literals describing how the action changes the world state. 

\subsection{Task Definition}
\label{sec:task_define}

The task is formally defined as: $\mathcal{M}: \mathcal{N} \rightarrow D, D \models \mathcal{N}$, where $\mathcal{M}$ is a mapping function (implemented by an LLM) that generates world model $D$ from the natural language description $\mathcal{N}$. $\models$ denotes semantic satisfaction.
Each $\mathcal{N}$ contains the following components: 
i) A general description describing the overall objective of the domain; 
ii) A set of predicates $\mathcal{N}_{F} = \{f_1, ..., f_n\}$ where each predicate is described with its signature (e.g., ``\textit{(conn ?x ?y)}'') and an explanation (e.g., ``\textit{Indicates a connection between two places ?x and ?y}''); 
iii) A set of actions $\mathcal{N}_{A} = \{a_1, ..., a_m\}$ where each action is described with: its signature (e.g., ``\textit{move <?curpos> <?nextpos>}'') and an explanation (e.g., ``\textit{Allows the robot to move from place <?curpos> to place <?nextpos>}'').
Note that to evaluate LLMs' inherent world modeling capabilities, action descriptions in $\mathcal{N}_{A}$ are intentionally kept at a high level, without explicit specifications of preconditions $\varphi$ and effects $\mathcal{E}$. This design choice allows us to assess how well LLMs can infer the underlying world dynamics and constraints from purely descriptive text.
A comparative analysis of model performance conditioned on different description styles is presented in Section~\ref{sec:concrete}.

\subsection{Evaluation Metrics}

We directly evaluate generated world models, addressing the ambiguity associated with indirect evaluations~\cite{guan2023leveraging,dainese2024generating}. 
In addition, we propose using execution-based metrics, overcoming the randomness of LLM-based evaluation~\cite{wang2023bytesized32}.
Specifically, we established the following evaluation metrics: 
\textit{(i)} \textit{\textbf{Executability (\exec):}} Measures whether the generated PDDL can be successfully parsed and validated by standard PDDL validators. 
\textit{(ii)} \textit{\textbf{Structural Similarity (\simmetric):}} Quantifies the textual similarity between the generated and ground truth PDDL using normalized Levenshtein ratio. 
\textit{(iii)} \textit{\textbf{Component-wise F1 Scores:}} When generated PDDL achieves executability (\exec = 1), we perform fine-grained analysis by calculating the macro-averaged F1 score for each component type (predicates, actions, etc.). More specifically, we compute F1 scores for predicates (\textbf{\fonepred}), parameters (\textbf{\foneparam}), preconditions (\textbf{\foneprecond}), and effects (\textbf{\foneeff}) by parsing both generated and ground truth PDDL into structured representations. 

\section{Benchmark Construction}

\begin{figure*}[htbp]
    \begin{minipage}{0.77\linewidth}  
        \centering
        \includegraphics[width=\linewidth]{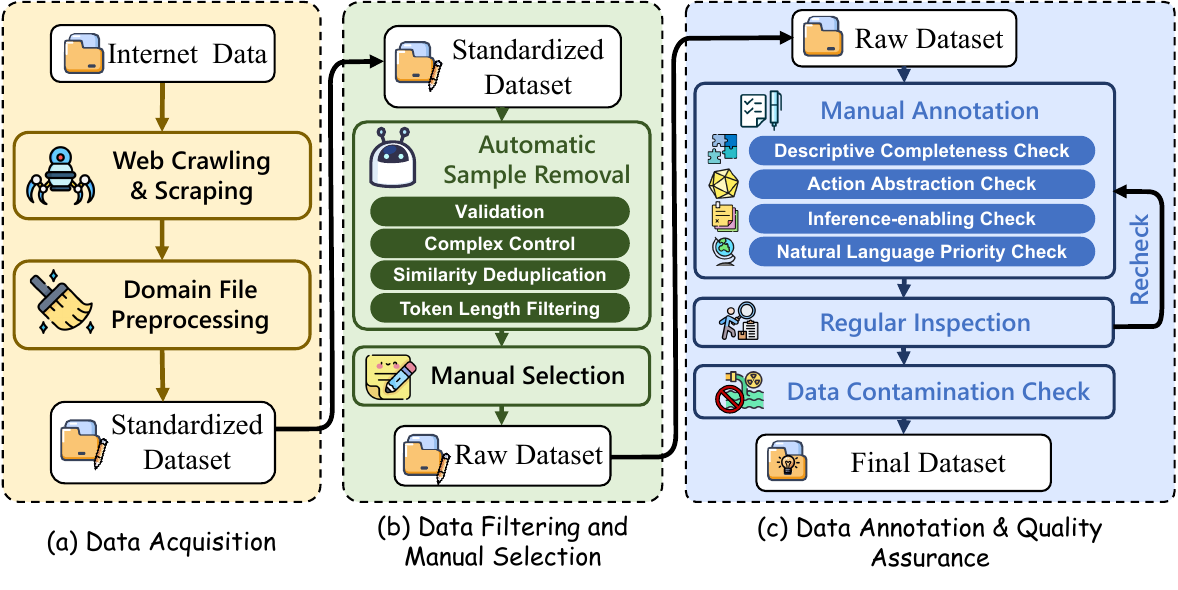}
    \end{minipage}%
    \hfill
    \begin{minipage}{0.23\linewidth}  
        \centering
        \resizebox{\linewidth}{!}{  
            \begin{tabular}{lcc}
                \toprule
                \textbf{Statistic} & \textbf{Number} \\
                \midrule
                \rowcolor[HTML]{F2F2F2} 
                \textit{Domain Count} &  \\
                \midrule
                Domain & 103 \\
                - Train & 2 \\
                - Test & 101 \\
                \midrule
                \rowcolor[HTML]{F2F2F2} 
                \textit{Token Count} &  \\
                \midrule
                Description & 851.6 $\pm$ 515.2 \\
                - Min/Max & [159, 2814] \\
                Domain & 1187.2 $\pm$ 1212.1 \\
                - Min/Max & [85, 7514] \\
                \midrule
                \rowcolor[HTML]{F2F2F2} 
                \textit{Line Count} &  \\
                \midrule
                Domain & 75.4 $\pm$ 62.9 \\
                - Min/Max & [9, 394] \\
                \midrule
                \rowcolor[HTML]{F2F2F2} 
                \textit{Component Count} &  \\
                \midrule
                Actions & 4.5 $\pm$ 2.8 \\
                - Min/Max & [1, 16] \\
                Predicates & 8.1 $\pm$ 4.8 \\
                - Min/Max & [1, 25] \\
                Types & 1.1 $\pm$ 1.3 \\
                - Min/Max & [1, 8] \\
                \bottomrule
            \end{tabular}
        }
    \end{minipage}
    \caption{\textit{Left}: Dataset construction process including: (a) \textit{Data Acquisition} (\S\ref{sec:data_acquisition}); (b) \textit{Data Filtering and Manual Selection} (\S\ref{sec:data_filtering}); (c) \textit{Data Annotation and Quality Assurance} (\S\ref{sec:data_annotation} and \S\ref{sec:quality_assurance}). \textit{Right}: Key statistics of \texttt{\benchmark}. Tokens are counted by GPT-2~\cite{openai2019gpt2} tokenizer. The style is referenced from \citet{chen-etal-2024-m3cot}.}
    \label{fig:combined}
\end{figure*}

The overall process of benchmark construction is shown in Figure~\ref{fig:combined}. In this section, we provide a detailed explanation of each stage.

\subsection{Data Acquisition}
\label{sec:data_acquisition}

Our benchmark construction process began with collecting PDDL files from various public repositories and planning competitions. 
Through this initial collection phase, we accumulated 1,801 raw PDDL files.
We performed several preprocessing steps to standardize the data format (e.g., convert files with BOM encoding to standard UTF-8).
The processed files served as the foundation for our dataset construction.

\subsection{Data Filtering and Manual Selection}
\label{sec:data_filtering}

To ensure the quality and reliability of \benchmark, we implemented a comprehensive filtering pipeline: 
\textit{(i)} \noindent \textit{\textbf{Validation:}} 
We employed a PDDL domain parser to perform syntax validation on each file;
\textit{(ii)} \noindent \textit{\textbf{Similarity Deduplication:}} 
We eliminated duplicate entries by computing pairwise cosine similarity on TF-IDF vectorized PDDL content, removing files with similarity scores exceeding 0.9;
\textit{(iii)} \noindent \textit{\textbf{Complexity Control:}} 
We removed domains with over 40 predicates or 20 actions to balance expressiveness with practical utility.
\textit{(iv)} \noindent \textit{\textbf{Token Length Filtering:}} We removed files exceeding 5,000 tokens using GPT-2~\cite{openai2019gpt2} tokenizer to ensure compatibility with model context windows.
Additionally, we conducted manual selection to eliminate domains that were not designed for world modeling (such as blocksworld-mystery) and low-quality cases that were not captured by the automated filtering methods.
After this process, we obtained 264 high-quality PDDL domain specifications.


\subsection{Data Annotation}
\label{sec:data_annotation}

After obtaining the high-quality PDDL domains, we manually annotated natural language descriptions for each domain. 
To ensure the quality of annotations, we recruited 6 computer science graduates as annotators.
The annotated description followed the structured format described in Section~\ref{sec:task_define}, and annotators were required to follow the annotation criteria:
\textit{(i)} \textit{\textbf{Descriptive Completeness:}} Annotations must contain all required components;
\textit{(ii)} \textit{\textbf{Action Abstraction:}} Action descriptions should avoid explicit references to formal preconditions and effects;
\textit{(iii)} \textit{\textbf{Inference-Enabling:}} Descriptions should contain sufficient contextual information to allow models to infer the underlying dynamics;
\textit{(iv)} \textit{\textbf{Natural Language Priority:}} Technical terminology should be minimized in favor of natural language explanations.
Examples of \benchmark can be found in Appendix~\ref{app:dataset-example}.

\subsection{Quality Assurance}
\label{sec:quality_assurance}

\noindent \textbf{Manual Recheck}
To maintain rigorous quality standards throughout the annotation process, we established a review system supervised by two senior experts. 
These experts conducted regular inspections of the annotations, ensuring accuracy and consistency. 
Inspectors must verify all data twice to determine if the annotated examples meet the specified annotation standards. Examples are accepted only if both inspectors approve them. The verification results showed "almost perfect agreement" with a Fleiss Kappa~\cite{10.2307/2529310} score of 0.82.
Through this comprehensive quality control process, we compiled a final curated dataset of 103 domains with gold-standard descriptions. 

\begin{figure}[t]
    \centering
    \includegraphics[width=0.8\linewidth]{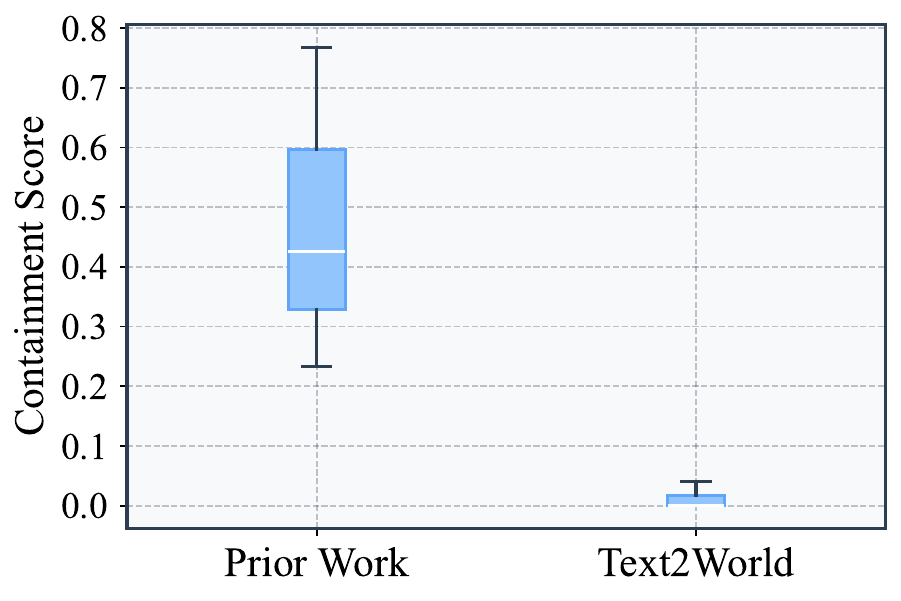}
    \caption{
    n-gram contamination rate of \benchmark and prior works.
    }
    \label{fig:cont}
\end{figure}

\noindent \textbf{Data Contamination}
As shown by~\citet{carlini2021extracting}, LLMs can memorize training data rather than truly model the world. 
To assess potential contamination between LLMs' training data and \benchmark, we generated complete PDDL domains from the first 20 tokens using GPT-4~\cite{openai2023gpt4} and calculated contamination rates based on tokenized 10-grams with up to 4 mismatches~\cite{touvron2023llama}, excluding PDDL-specific keywords and variables. 
We also compared these results with previous studies~\cite{guan2023leveraging,smirnov2024generating}. 
Figure~\ref{fig:cont} shows that \benchmark has a lower contamination rate ($\mu = 0.04$ vs. $\mu = 0.47$), suggesting its performance reflects domain understanding rather than memorization. However, the complete elimination of contamination remains challenging due to PDDL's widespread use.

\subsection{Data Analysis}

This section provides some detailed data analysis to better understand \benchmark.

\noindent \textbf{Core Statistics} We designated 2 domains as in-context exemplars (train set), with the remaining 101 samples forming our test set.

\begin{figure}[t]
    \begin{minipage}{\columnwidth}  
        \centering
        \includegraphics[width=\linewidth]{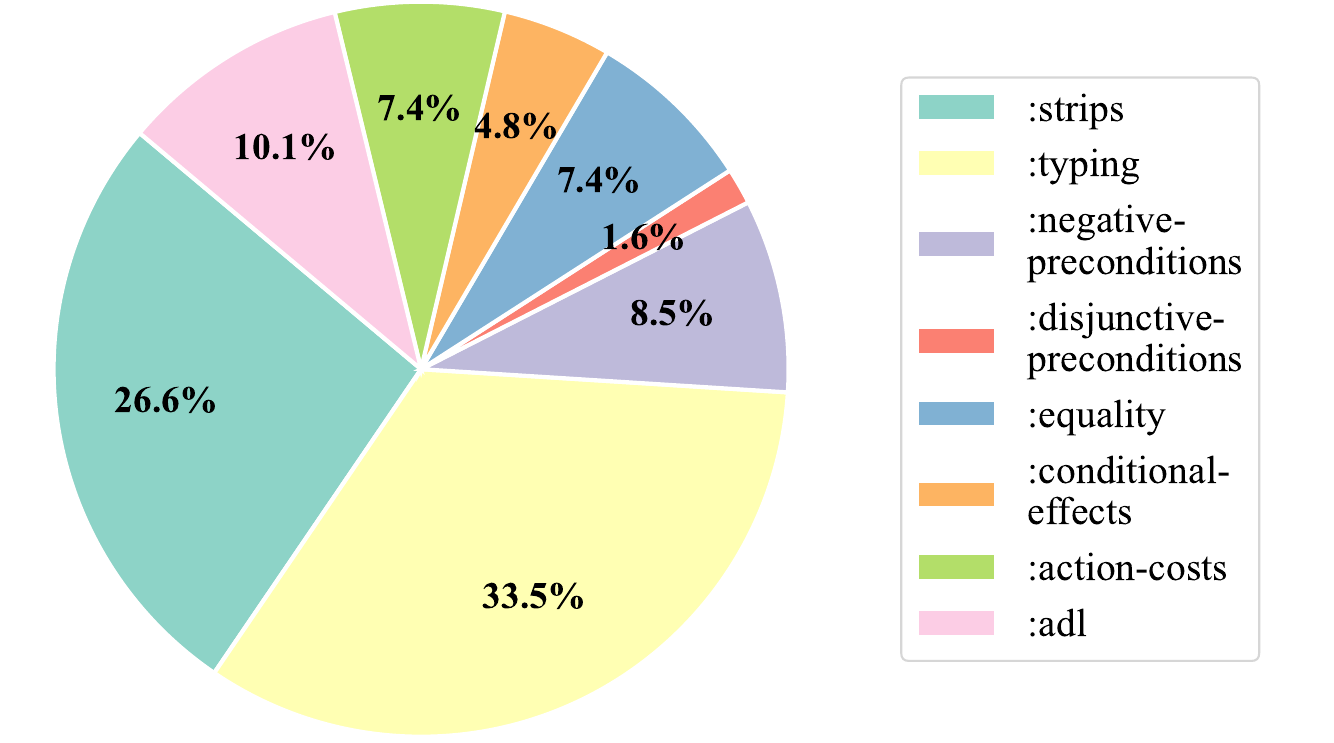}
    \end{minipage}
    
    \begin{minipage}{\columnwidth}
        \centering
        \includegraphics[width=\linewidth]{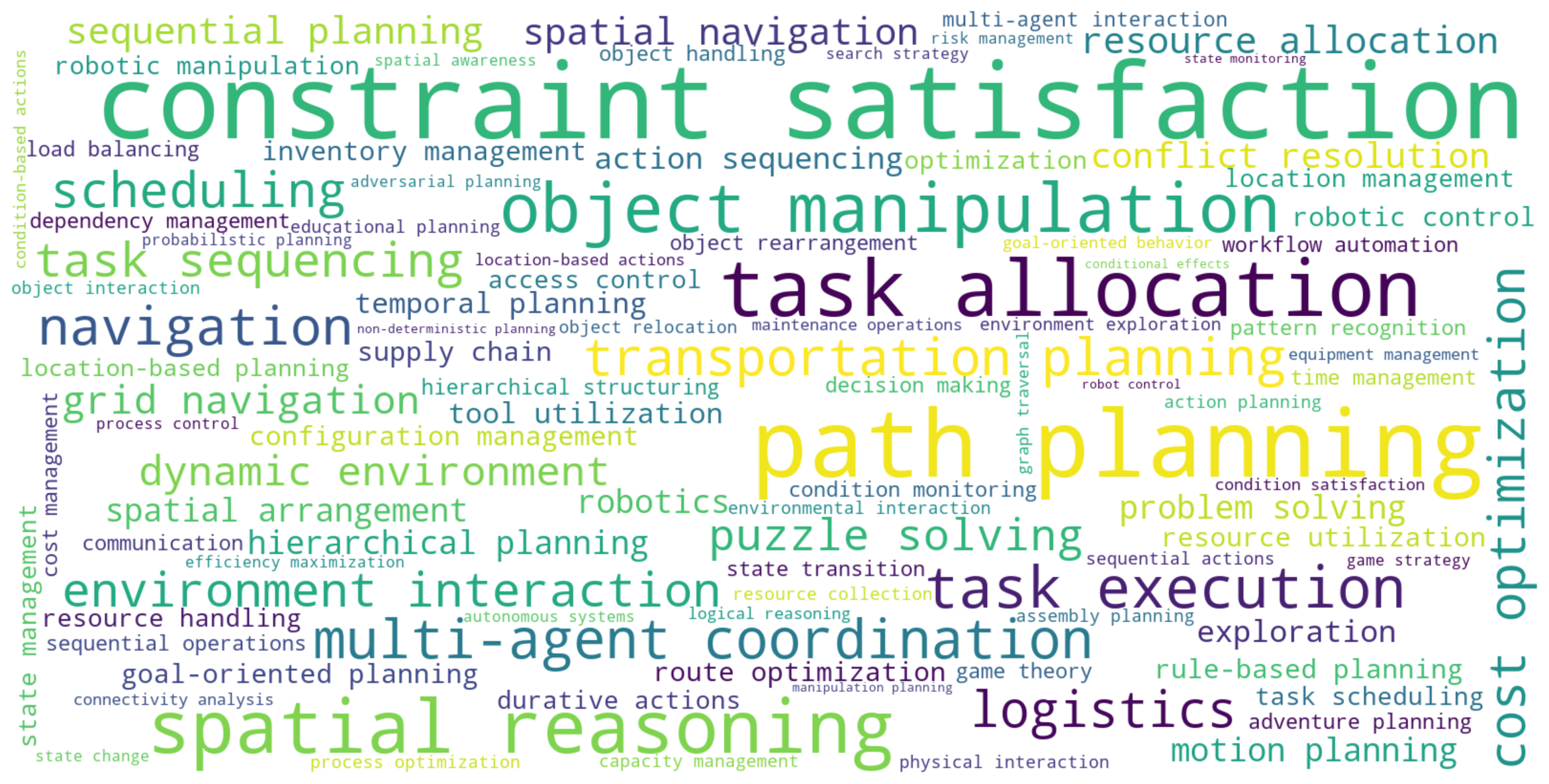}
    \end{minipage}
    
    \caption{
    \textit{Top}: The frequency of requirements distribution. \textit{Bottom}: Word cloud of concepts in \benchmark.}
    \label{fig:three_plots}
\end{figure}

\input{table/statistics}
\noindent \textbf{Semantic Analysis} 
\label{sec:wordcloud}
We use LLMs to extract high-level domain characteristics to better understand the conceptual distribution of \benchmark,
As shown in Figure~\ref{fig:three_plots} (Bottom), common themes such as \textit{path planning}, \textit{constraint satisfaction}, and \textit{task allocation}, among others, emerge. 


\noindent \textbf{Requirements Analysis} 
\label{sec:requirements_analysis}
A PDDL requirement specifies a formal capability needed to express a domain, often reflecting its complexity. For instance, \texttt{:typing} stands for allowing the usage of typing for objects.
As shown in Figure~\ref{fig:three_plots} (Top), there are eight different requirement type in \benchmark. We also provide an in-depth analysis of requirement type in Appendix~\ref{app:detailed_data_analysis}.

\subsection{Preliminary Experiment}
\label{sec:llmself}
In previous works, LLMs have been employed to evaluate the action dynamics of world models generated by LLMs themselves~\cite{wang2023bytesized32}. 
To further assess the ability of LLMs to detect errors in world models, we conducted a preliminary experiment where we first used \texttt{claude-3.5-sonnect} for \benchmark. Subsequently, human annotators and the LLM independently evaluated the generated action dynamics to identify potential errors.
The inter-annotator agreement between human ratings and LLM ratings, measured using Cohen’s $\kappa$, was 0.10, indicating a low level of agreement. This suggests that predicting the correctness of PDDL domains using an LLM is particularly challenging, highlighting the need for more discriminative evaluation metrics.
Prompting examples and more results can be found in Appendix~\ref{app:self_eval}.

\section{Experiments}

\input{section/experiment}
\begin{table*}[htbp]
\centering
\caption{Performance comparison of different LLMs on \benchmark. EC\textsubscript{k} denotes the setting where models are allowed k correction attempts (EC\textsubscript{0}: zero-shot without correction, EC\textsubscript{3}: with 3 correction attempts).}
\label{tab:main}
\resizebox{\textwidth}{!}{%
\begin{tabular}{l|c|cc|cc|cc|cc|cc|cc}
\toprule
\multirow{2}{*}{\textbf{Model Family}} & \multirow{2}{*}{\textbf{Version}} & \multicolumn{2}{c|}{\textbf{\exec} $\uparrow$} & \multicolumn{2}{c|}{\textbf{\simmetric} $\uparrow$} & \multicolumn{2}{c|}{\textbf{\fonepred} $\uparrow$} & \multicolumn{2}{c|}{\textbf{\foneparam} $\uparrow$} & \multicolumn{2}{c|}{\textbf{\foneprecond} $\uparrow$} & \multicolumn{2}{c}{\textbf{\foneeff} $\uparrow$} \\
\cmidrule{3-14} & & \textbf{EC\textsubscript{0}} & \textbf{EC\textsubscript{3}} & \textbf{EC\textsubscript{0}} & \textbf{EC\textsubscript{3}} & \textbf{EC\textsubscript{0}} & \textbf{EC\textsubscript{3}} & \textbf{EC\textsubscript{0}} & \textbf{EC\textsubscript{3}} & \textbf{EC\textsubscript{0}} & \textbf{EC\textsubscript{3}} & \textbf{EC\textsubscript{0}} & \textbf{EC\textsubscript{3}} \\
\midrule
\multirow{1}{*}{\textsc{OpenAI-o1}}  & {\small\texttt{o1-mini}} & 49.5 & 69.3 & 82.5 & 82.2 & 48.4 & 66.3 & 36.4 & 49.7 & 28.9 & 38.0 & 31.7 & 42.1 \\
\midrule
\textsc{OpenAI-o3} & {\small\texttt{o3-mini}} & 54.5 & 84.2 & 83.0 & 81.9 & 53.9 & 81.1 & 43.7 & 63.0 & 36.8 & 50.4 & 39.4 & 53.8 \\
\midrule
\multirow{2}{*}{\textsc{GPT-4}} & {\small\texttt{gpt-4o}} & 60.4 & 75.2 & 84.5 & 84.1 & 59.6 & 72.1 & 56.5 & 68.1 & 49.3 & 56.4 & 47.8 & 56.7 \\
 & {\small\texttt{gpt-4o-mini}} & 48.5 & 72.3 & 82.6 & 82.2 & 48.1 & 70.1 & 47.1 & 67.3 & 34.9 & 47.5 & 38.2 & 52.7 \\
\midrule
\multirow{1}{*}{\textsc{GPT-3.5}}  & {\small\texttt{turbo-0125}} & 41.6 & 56.4 & 81.9 & 81.6 & 41.2 & 55.8 & 39.6 & 53.8 & 30.2 & 39.2 & 27.5 & 37.7 \\
\midrule
\textsc{Claude-3.5} & {\small\texttt{sonnet}} & 45.5 & 64.4 & 73.2 & 66.8 & 45.5 & 62.5 & 41.5 & 48.8 & 37.4 & 44.0 & 38.4 & 45.0 \\
\midrule
\multirow{2}{*}{\textsc{LLaMA-2}} & {\small\texttt{7b-instruct}} & 0.0 & 0.0 & 45.5 & 33.9 & 0.0 & 0.0 & 0.0 & 0.0 & 0.0 & 0.0 & 0.0 & 0.0 \\
 & {\small\texttt{70b-instruct}} & 0.0 & 0.0 & 48.7 & 48.6 & 0.0 & 0.0 & 0.0 & 0.0 & 0.0 & 0.0 & 0.0 & 0.0 \\
\midrule
\multirow{2}{*}{\textsc{LLaMA-3.1}} & {\small\texttt{8b-instruct}} & 0.0 & 0.0 & 74.3 & 74.9 & 0.0 & 0.0 & 0.0 & 0.0 & 0.0 & 0.0 & 0.0 & 0.0 \\
 & {\small\texttt{70b-instruct}} & 0.0 & 0.0 & 83.6 & 79.2 & 0.0 & 0.0 & 0.0 & 0.0 & 0.0 & 0.0 & 0.0 & 0.0 \\
\midrule
\multirow{2}{*}{\textsc{DeepSeek}} & {\small\texttt{deepseek-v3}} & 56.4 & 79.2 & \cellcolor[HTML]{F2F2F2}\textbf{84.7} & \cellcolor[HTML]{F2F2F2}\textbf{84.2} & 55.9 & 75.6 & 53.7 & 74.4 & 45.1 & 58.6 & 46.7 & 61.5 \\ 
& {\small\texttt{deepseek-r1}} & \cellcolor[HTML]{F2F2F2}\textbf{72.3} & \cellcolor[HTML]{F2F2F2}\textbf{89.1} & 84.3 & 84.0 & \cellcolor[HTML]{F2F2F2}\textbf{71.7} & \cellcolor[HTML]{F2F2F2}\textbf{86.7} & \cellcolor[HTML]{F2F2F2}\textbf{64.0} & \cellcolor[HTML]{F2F2F2}\textbf{76.3} & \cellcolor[HTML]{F2F2F2}\textbf{57.6} & \cellcolor[HTML]{F2F2F2}\textbf{65.0} & \cellcolor[HTML]{F2F2F2}\textbf{58.8} & \cellcolor[HTML]{F2F2F2}\textbf{67.3} \\ 
\midrule
\multirow{4}{*}{\textsc{CodeLLaMA}} & {\small\texttt{7b-instruct}} & 17.8 & 22.8 & 60.2 & 57.6 & 17.8 & 18.8 & 17.2 & 18.2 & 11.3 & 12.2 & 10.7 & 11.1 \\
 & {\small\texttt{13b-instruct}} & 7.9 & 8.9 & 57.6 & 55.0 & 7.9 & 8.9 & 7.9 & 8.9 & 4.9 & 5.9 & 5.2 & 6.1 \\
 & {\small\texttt{34b-instruct}} & 7.9 & 8.9 & 34.2 & 7.6 & 7.9 & 8.6 & 7.9 & 8.4 & 5.0 & 5.0 & 5.4 & 5.4 \\
 & {\small\texttt{70b-instruct}} & 16.8 & 16.8 & 54.0 & 14.0 & 16.4 & 16.4 & 16.8 & 16.8 & 10.7 & 10.7 & 14.1 & 14.1 \\
\bottomrule
\end{tabular}%
}
\end{table*}

\subsection{Experimental Setup}


We evaluate several state-of-the-art LLMs, including \textit{GPT-4}~\cite{openai2023gpt4}, \textit{GPT-3.5}~\cite{opeiai2022gpt}, \textit{Claude-3.5}~\cite{claude_3.5_sonnet}, and \textit{LLaMA-3.1}~\cite{meta_llama_3_1}, \textit{DeepSeek-v3}~\cite{liu2024deepseek}, \textit{CodeLlaMA}~\cite{roziere2023code}, \textit{LlaMA-2}~\cite{touvron2023llama}, etc.
We also evaluated Large Reasoning Models (LRMs) trained using reinforcement learning, such as \textit{DeepSeek-R1}~\cite{deepseekai2025deepseekr1incentivizingreasoningcapability}, \textit{OpenAI-o1}~\cite{openai2024o1} and \textit{OpenAI-o3}~\cite{openai2025o3}.
We set temperature = 0 for each model for all experiments to maintain reproducibility.
We employ tarski~\footnote{\url{https://github.com/aig-upf/tarski}} library to check syntactic correctness and executability.
We prompt LLMs to generate symbolic world models under a zero-shot setting with chain-of-thought reasoning~\cite{wei2022chain}. 
In error-correction experiments, LLMs refine outputs based on validator-reported syntax errors, denoted as $\text{EC}_{3}$ for $k$ attempts.
Evaluation of open-sourced models were conducted on NVIDIA A100 GPUs with 80GB memory.
We access proprietary models through their official API platform.
Prompt examples can be found in Appendix~\ref{app:prompt}.


\subsection{Experimental Results}
\label{sec:main_exp}

Several conclusions can be drawn from Table~\ref{tab:main}:
\textit{(i)} \textbf{\textit{The most advanced LLMs still struggle with \benchmark.}}
For example, the best-performing model, \textit{DeepSeek-R1}, achieves F1 scores below 60\% for both preconditions (\foneprecond) and effects (\foneeff) under the without error correction setting.
This highlights the limitations of current LLMs in world modeling tasks.
\textit{(ii)} \textit{\textbf{Large reasoning models trained with reinforcement learning exhibit superior world modeling capabilities.}} These models, such as \textit{DeepSeek-R1}~\cite{deepseekai2025deepseekr1incentivizingreasoningcapability}, outperform others in executability, structural similarity, and component-wise performance, indicating that RL-based training enhances the ability of models to generate structured and valid world models.
\textit{(iii)} \textit{\textbf{The ability of models to benefit from error correction is evident.}} For instance, \textit{GPT-4} (\texttt{gpt-4o-mini}) demonstrates a notable improvement in executability, increasing from 48.5\% to 72.3\% after three correction attempts. 

\section{Analysis}

\subsection{Statistical Analysis}
\label{sec:stat_ana}

We conducted a one-way ANOVA~\cite{girden1992anova} to evaluate the impact of correction attempts on model performance, excluding anomalous zero values. The results showed a significant improvement with three correction attempts ($F = 27.48, p = 0.00012$), indicating that correction attempts lead to a notable enhancement in model performance.

\subsection{Error Analysis}
\label{sec:analysis}

The interpretable nature of generating symbolic world models can be utilized for a deeper manual analysis of the failure modes. 
We select the results from \texttt{claude-3.5-sonnect} under the few-shot setting for manual error analysis. 
Errors are categorized into syntax and semantic errors, where syntax errors occur when the generated domain cannot be validated ($\exec=0$), and semantic errors arise when the generated world model does not align with action dynamics or fails to follow the natural language description.
The distribution for each error type and detailed explanations are presented in Appendix~\ref{app:analysis_detail}.

\begin{figure*}[htbp]
    \centering
    \includegraphics[width=\linewidth]{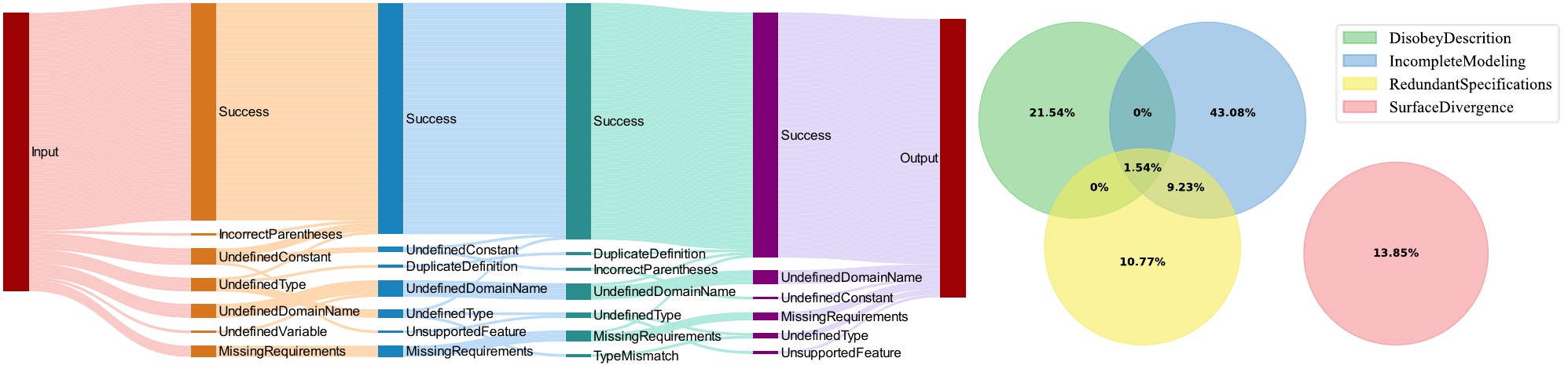}
    \caption{\textit{Left}: The distribution of syntax error types during the progression of correction. \textit{Right}: The distribution of semantic error types.}
    \label{fig:sankey_venn}
\end{figure*}

\noindent \textbf{Syntax Errors}
Figure~\ref{fig:sankey_venn} (\textit{Left}) shows the distribution of syntax errors during correction. 
Common errors like \texttt{UndefinedConstant} and \texttt{IncorrectParentheses} decrease over correction steps, indicating improvements in syntax validation, though errors like \texttt{UndefinedDomainName} and \texttt{UndefinedType} persist.

\noindent \textbf{Semantic Error}
Figure~\ref{fig:sankey_venn} (\textit{Right}) illustrates the distribution of semantic errors. 
Semantic errors are categorized into four types: 
\textit{(i)} \texttt{DisobeyDescription} involves direct violations of descriptions.
\textit{(ii)} \texttt{IncompleteModeling}, where the world model lacks necessary components. 
\textit{(iii)} \texttt{RedundantSpecifications} refers to superfluous preconditions or effects;
and \textit{(iv)} \texttt{SurfaceDivergence} involves surface-level variations that preserve semantic equivalence to gold domain.
In addition, since a domain may encompass various action dynamics, different error types can occur simultaneously. 
For instance, nearly 10\% of cases exhibited both \texttt{IncompleteModeling} and \texttt{RedundantSpecifications} concurrently.


\vspace{-5pt}

\section{Exploration}

In addition to the zero-shot CoT evaluation in Section~\ref{sec:main_exp}, we further evaluate the models on \benchmark with five different strategies: (1) \textit{Test-time Scaling}; (2) \textit{In-Context Learning}; (3) \textit{Fine-tuning}; (4) \textit{Agent Training}; (5) \textit{Inference with Concrete Description}.

\subsection{Test-time Scaling}
\label{sec:more_correction}
\begin{figure}[t]
    \centering
    \includegraphics[width=\linewidth]{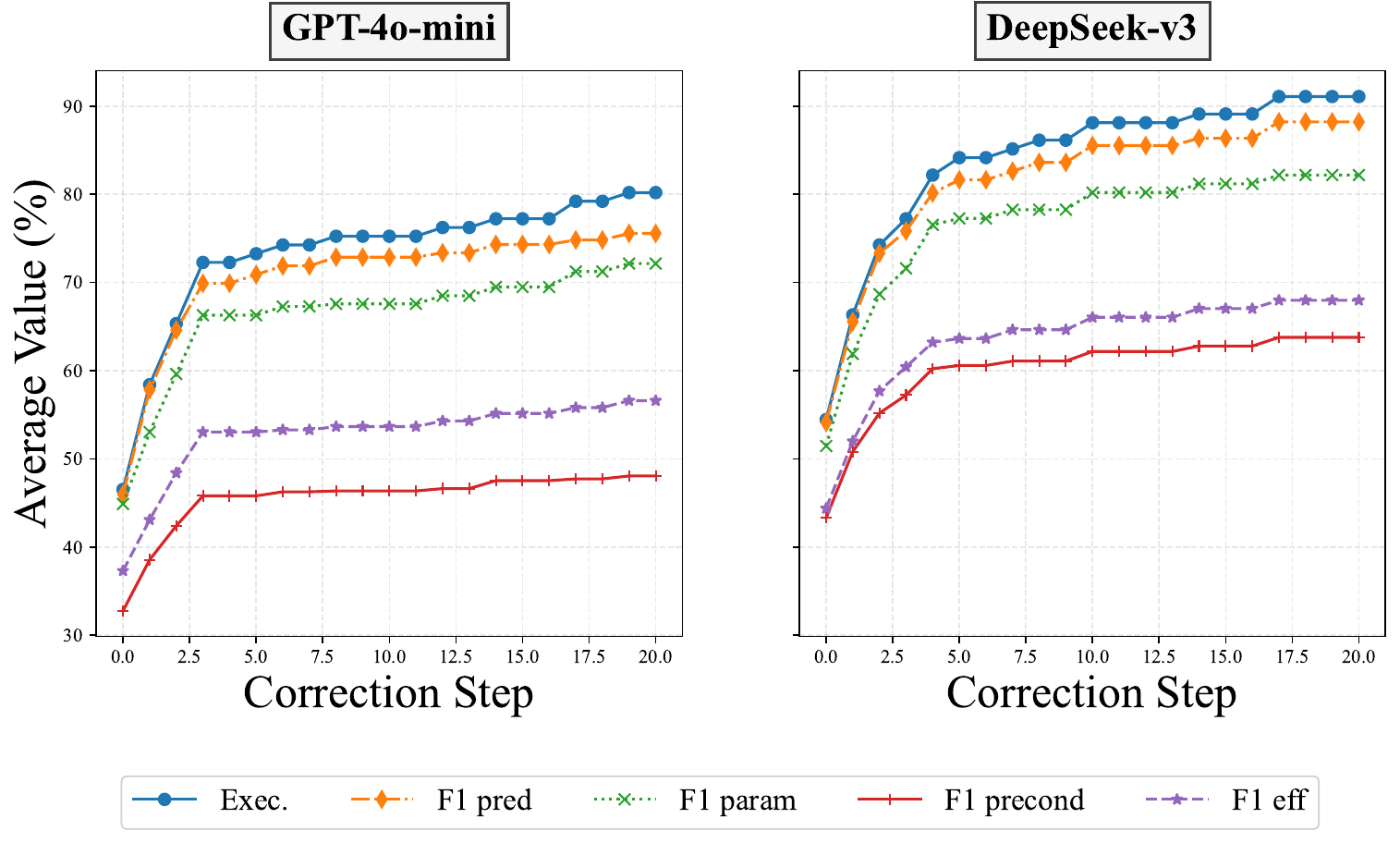}
    \caption{
    The performance of \texttt{gpt-4o-mini} (left) and \texttt{deepseek-v3} (right) under different test-time compute budgets, showing consistent improvement with increased compute.
    }
    \label{fig:20correction}
\end{figure}

Recently, test-time scaling has demonstrated remarkable potential~\cite{openai2024o1,deepseekai2025deepseekr1incentivizingreasoningcapability}.
We use the error information from the syntax parser as feedback and assess whether increasing the test-time compute budget can enhance the LLM's performance.
As shown in Figure~\ref{fig:20correction}, the model exhibits consistent improvement with increased test-time computation.
More advanced test-time scaling strategies may serve as a viable approach to enhancing the model's world modeling ability~\cite{chen2025ecm}.

\vspace{-2pt}

\subsection{In-Context Learning} 
\label{sec:few-shot}

We also perform a few-shot evaluation in Section~\ref{sec:few-shot}, where we carefully select demonstration ``\textit{gripper}'' and ``\textit{blocks}'' that are structurally similar but semantically distinct from the test cases to prevent data leakage.
As shown in Table~\ref{tab:optimization}, we observe that different models exhibit varying degrees of improvement from in-context learning. For instance, \texttt{claude-3.5-sonnect} demonstrates a substantial enhancement, achieving over a 20\% increase in the component-wise F1 score. However, for \texttt{gpt-4o-mini}, incorporating few-shot examples resulted in a decrease in model performance.

\subsection{Fine-tuning} 
\label{sec:finetuning}

We leverage the AgentGen~\cite{hu2024agentgen} framework to synthesize 601 PDDL domains and their corresponding descriptions for fine-tuning \textit{LLaMA-3.1}~\cite{meta_llama_3_1} to investigate potential improvements in their world modeling capabilities.
As shown in Table~\ref{tab:optimization}, fine-tuning can lead to significant improvements in model performance. For instance, the fine-tuned Llama-3.1-70B demonstrated performance comparable to GPT-4o-mini, highlighting that supervised fine-tuning is an effective method for bridging the gap between open-source and proprietary models.
Moreover, larger models tend to benefit more from supervised fine-tuning, with the 70B LLaMA-3.1 showing greater improvement than the 8B model.

\subsection{Agent Training}
\label{sec:agent_training}
Many studies have demonstrated that supervised fine-tuning on agent trajectories can enhance a model's performance on agentic tasks~\cite{hu2024agentgen,zeng2023agenttuning} (i.e., agent training). 
Some previous works also discussed that a good agent model requires a sufficiently strong internal world representation~\cite{lecun2022path}. 
Therefore, in this section, we explore whether agent training can improve the model's world modeling capabilities. 
More specifically, we trained \textit{LLaMA-2-70B} model on AgentInstruct~\cite{zeng2023agenttuning}. 
As shown in Table~\ref{tab:optimization}, the model's world modeling capabilities are enhanced post-agent training, indicating a positive correlation between performance on agentic tasks and the model's world modeling abilities.

\begin{table*}[htbp]
\centering
\caption{The experimental results of models under different settings: (1) In-context learning (\S\ref{sec:few-shot}); (2) Fine-tuning, and fine-tuning with LoRA~\cite{hu2021lora} (\S\ref{sec:finetuning}); (3) Agent training (\S\ref{sec:agent_training}).}
\label{tab:optimization}
\resizebox{\textwidth}{!}{%
\begin{tabular}{l|ll|ll|ll|ll|ll|ll}
\toprule
\multirow{1}{*}{\textbf{Model Family}} & \multicolumn{2}{c|}{\textbf{\exec}} & \multicolumn{2}{c|}{\textbf{\simmetric}} & \multicolumn{2}{c|}{\textbf{\fonepred}} & \multicolumn{2}{c|}{\textbf{\foneparam}} & \multicolumn{2}{c|}{\textbf{\foneprecond}} & \multicolumn{2}{c}{\textbf{\foneeff}} \\
\cmidrule{2-13} & \textbf{EC\textsubscript{0}} & \textbf{EC\textsubscript{3}} & \textbf{EC\textsubscript{0}} & \textbf{EC\textsubscript{3}} & \textbf{EC\textsubscript{0}} & \textbf{EC\textsubscript{3}} & \textbf{EC\textsubscript{0}} & \textbf{EC\textsubscript{3}} & \textbf{EC\textsubscript{0}} & \textbf{EC\textsubscript{3}} & \textbf{EC\textsubscript{0}} & \textbf{EC\textsubscript{3}} \\
\midrule
\rowcolor[HTML]{F2F2F2} 
\multicolumn{13}{c}{\textit{In-Context Learning}} \\
\midrule
\textsc{Claude-3.5-sonnet} & 45.5 & 64.4 & 73.2 & 66.8 & 45.5 & 62.5 & 41.5 & 48.8 & 37.4 & 44.0 & 38.4 & 45.0 \\
~$w.$ \textsc{2-shot} & 
78.2\textsubscript{\textcolor{darkgreen}{+32.7}} & 88.1\textsubscript{\textcolor{darkgreen}{+23.7}} & 83.9\textsubscript{\textcolor{darkgreen}{+10.7}} & 82.3\textsubscript{\textcolor{darkgreen}{+15.5}} & 77.0\textsubscript{\textcolor{darkgreen}{+31.5}} & 86.1\textsubscript{\textcolor{darkgreen}{+23.6}} & 75.2\textsubscript{\textcolor{darkgreen}{+33.7}} & 82.1\textsubscript{\textcolor{darkgreen}{+33.3}} & 65.6\textsubscript{\textcolor{darkgreen}{+28.2}} & 71.3\textsubscript{\textcolor{darkgreen}{+27.3}} & 67.2\textsubscript{\textcolor{darkgreen}{+28.8}} & 73.4\textsubscript{\textcolor{darkgreen}{+28.4}} \\
\midrule
\textsc{Deepseek-r1} & 72.3 & 89.1 & 84.3 & 84.0 & 71.7 & 86.7 & 64.0 & 76.3 & 57.6 & 65.0 & 58.8 & 67.3 \\ 
~$w.$ \textsc{2-shot} & 69.3\textsubscript{\textcolor{red}{-3.0}} & 90.1\textsubscript{\textcolor{darkgreen}{+1.0}} & 83.8\textsubscript{\textcolor{red}{-0.5}} & 83.5\textsubscript{\textcolor{red}{-0.5}} & 68.4\textsubscript{\textcolor{red}{-3.3}} & 87.7\textsubscript{\textcolor{darkgreen}{+1.0}} & 64.6\textsubscript{\textcolor{darkgreen}{+0.6}} & 79.1\textsubscript{\textcolor{darkgreen}{+2.8}} & 56.0\textsubscript{\textcolor{red}{-1.6}} & 66.9\textsubscript{\textcolor{darkgreen}{+1.9}} & 57.6\textsubscript{\textcolor{red}{-1.2}} & 68.9\textsubscript{\textcolor{darkgreen}{+1.6}} \\
\midrule
\textsc{GPT-4o-mini}  & 48.5 & 72.3 & 82.6 & 82.2 & 48.1 & 70.1 & 47.1 & 67.3 & 34.9 & 47.5 & 38.2 & 52.7 \\
~$w.$ \textsc{2-shot} & 40.6\textsubscript{\textcolor{red}{-7.9}} & 69.3\textsubscript{\textcolor{red}{-3}} & 82.9\textsubscript{\textcolor{darkgreen}{+0.3}} & 82.4\textsubscript{\textcolor{darkgreen}{+0.2}} & 40.3\textsubscript{\textcolor{red}{-7.8}} & 67.2\textsubscript{\textcolor{red}{-2.9}} & 40.1\textsubscript{\textcolor{red}{-7}} & 67.0\textsubscript{\textcolor{red}{-0.3}} & 31.6\textsubscript{\textcolor{red}{-3.3}} & 49.3\textsubscript{\textcolor{darkgreen}{+1.8}} & 32.5\textsubscript{\textcolor{red}{-5.7}} & 54.8\textsubscript{\textcolor{darkgreen}{+2.1}} \\
\midrule
\rowcolor[HTML]{F2F2F2} 
\multicolumn{13}{c}{\textit{Fine-tuning (FT)}} \\
\midrule
\textsc{LLaMA-3.1-8B} & 0.0 & 0.0 & 74.3 & 74.9 & 0.0 & 0.0 & 0.0 & 0.0 & 0.0 & 0.0 & 0.0 & 0.0 \\
~$w.$ \textsc{FT} & 52.5\textsubscript{\textcolor{darkgreen}{+52.5}} & 68.3\textsubscript{\textcolor{darkgreen}{+68.3}} & 80.8\textsubscript{\textcolor{darkgreen}{+6.5}} & 80.6\textsubscript{\textcolor{darkgreen}{+5.7}} & 51.4\textsubscript{\textcolor{darkgreen}{+51.4}} & 65.4\textsubscript{\textcolor{darkgreen}{+65.4}} & 48.5\textsubscript{\textcolor{darkgreen}{+48.5}} & 60.6\textsubscript{\textcolor{darkgreen}{+60.6}} & 31.5\textsubscript{\textcolor{darkgreen}{+31.5}} & 38.1\textsubscript{\textcolor{darkgreen}{+38.1}} & 32.4\textsubscript{\textcolor{darkgreen}{+32.4}} & 40.2\textsubscript{\textcolor{darkgreen}{+40.2}} \\
\midrule
\textsc{LLaMA-3.1-70B} & 0.0 & 0.0 & 83.6 & 79.2 & 0.0 & 0.0 & 0.0 & 0.0 & 0.0 & 0.0 & 0.0 & 0.0 \\
~$w.$ \textsc{LoRA} & 48.5\textsubscript{\textcolor{darkgreen}{+48.5}} & 70.3\textsubscript{\textcolor{darkgreen}{+70.3}} & 83.8\textsubscript{\textcolor{darkgreen}{+0.2}} & 82.3\textsubscript{\textcolor{darkgreen}{+3.1}} & 47.9\textsubscript{\textcolor{darkgreen}{+47.9}} & 68.5\textsubscript{\textcolor{darkgreen}{+68.5}} & 48.5\textsubscript{\textcolor{darkgreen}{+48.5}} & 66.4\textsubscript{\textcolor{darkgreen}{+66.4}} & 39.9\textsubscript{\textcolor{darkgreen}{+39.9}} & 52.8\textsubscript{\textcolor{darkgreen}{+52.8}} & 40.6\textsubscript{\textcolor{darkgreen}{+40.6}} & 52.1\textsubscript{\textcolor{darkgreen}{+52.1}} \\
\midrule
\rowcolor[HTML]{F2F2F2} 
\multicolumn{13}{c}{\textit{Agent Training (AT)}} \\
\midrule
\textsc{LLaMA-2-70B} & 0.0 & 0.0 & 48.7 & 48.6 & 0.0 & 0.0 & 0.0 & 0.0 & 0.0 & 0.0 & 0.0 & 0.0 \\
\text{~$w.$ \textsc{AT}} & 7.9\textsubscript{\textcolor{darkgreen}{+7.9}} & 9.9\textsubscript{\textcolor{darkgreen}{+9.9}} & 65.6\textsubscript{\textcolor{darkgreen}{+16.9}} & 47.9\textsubscript{\textcolor{red}{-0.7}} & 7.3\textsubscript{\textcolor{darkgreen}{+7.3}} & 8.8\textsubscript{\textcolor{darkgreen}{+8.8}} & 7.3\textsubscript{\textcolor{darkgreen}{+7.3}} & 9.1\textsubscript{\textcolor{darkgreen}{+9.1}} & 6.1\textsubscript{\textcolor{darkgreen}{+6.1}} & 6.5\textsubscript{\textcolor{darkgreen}{+6.5}} & 5.7\textsubscript{\textcolor{darkgreen}{+5.7}} & 6.1\textsubscript{\textcolor{darkgreen}{+6.1}} \\
\bottomrule
\end{tabular}%
}
\end{table*}

\subsection{Inference with Concrete Description}
\label{sec:concrete}
\begin{figure}[t]
    \centering
    \includegraphics[width=\linewidth]{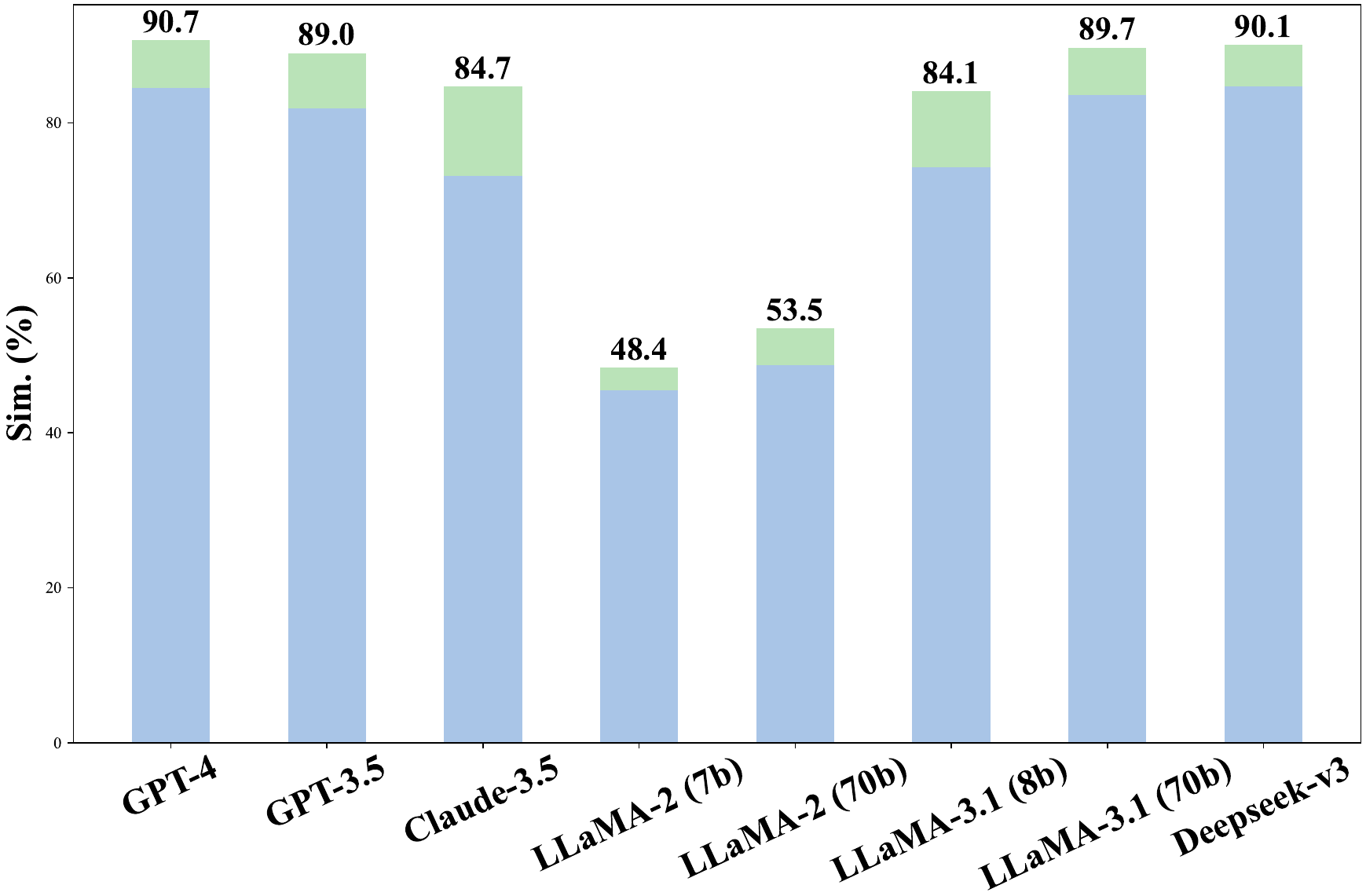}
    \caption{
    Comparison of model performance on abstract versus concrete domain descriptions, showing the base score for abstract descriptions (\textcolor[rgb]{0.663,0.773,0.906}{blue}) and the improvement gained from concrete descriptions (\textcolor[rgb]{0.663,0.867,0.655}{green}).
    }
    \label{fig:concrete}
\end{figure}

As is discussed in Section~\ref{sec:task_define}, we intentionally make the natural language description of a world model at a high level.
We refer to these high-level descriptions as "\textit{abstract descriptions}," in contrast to more detailed "\textit{concrete descriptions}" that explicitly specify preconditions and effects. 
Examples of both description types can be found in the Appendix~\ref{app:diffdesc}. 
Using concrete descriptions simplifies the task by requiring the model to directly map the provided text to a world specification, bypassing the need to infer symbolic action dynamics. 
The observed consistent improvement (as shown in Figure~\ref{fig:concrete}) supports the claim that the model's ability to deduce action dynamics from abstract descriptions is still lacking.
We also provide more detailed experimental results in Appendix~\ref{app:concrete_results}.

\vspace{-5pt}

\section{Related Work}

Neural world modeling is a long-standing research topic with widespread applications across various fields, including reinforcement learning~\citep{ha2018world, ha2018recurrent}, robotics~\citep{wu2023daydreamer}, and autonomous driving~\citep{guan2024world}, among others.
In recent years, LLMs trained on massive datasets have demonstrated zero-shot capabilities across a variety of tasks, including planning~\cite{zhao2023survey,qin2024large,huang2022language,hu2024hiagent}, robotics~\cite{mu2024robocodex,chen2024textbfemostextbfembodimentawareheterogeneoustextbfmultirobot}, analog design~\cite{lai2024analogcoder}, and more.
Preliminary studies propose directly using LLMs as world models~\citep{hao2023reasoning,wang2024can,wang2023promptagent,li2022emergent}, by taking the state and action as input and predicting the next state, but the unreliability and limited interpretability of LLM outputs can lead to accumulating errors.
Moreover, some studies have shown that autoregressive models perform poorly in predicting action effects~\cite{banerjee2020can,luo2023towards}.
Tree-planner~\cite{hu2023tree} instead proposes to constructing the possible action space using LLMs before executing.
Another line of work focuses on leveraging LLMs to construct symbolic world models~\citep{oswald2024large,silver2024generalized,smirnov2024generating,zhu2024language,wang2023bytesized32,wong2023word,vafa2024evaluating}. 
For example, ~\citet{guan2023leveraging} uses LLMs to generate a PDDL domain model and relies on human feedback to correct errors. 
AgentGen~\citep{hu2024agentgen} synthesizes diverse PDDL domains, aiming to create high-quality planning data. 
~\citet{xie2024making} propose to finetune LLMs for predicting precondition and effect of actions.
%
Despite the growing interest in this research direction, there is currently a lack of a comprehensive benchmark in this area.


\section{Conclusion}

We present \benchmark, a novel benchmark consisting of hundreds of domains designed to evaluate the world modeling capabilities of large language models (LLMs). 
Developed through a meticulous and thorough process, \benchmark provides a robust foundation for analysis. 
Additionally, we conducted an extensive evaluation involving 16 different LLMs from 9 model families based on \benchmark.
We hope that \benchmark will inspire future research in leveraging LLMs as world models.

\section*{Ethical Considerations}

\textbf{Data Access.} We collected the \benchmark data from open-source repositories and ensured that these repositories are available for academic research in accordance with our commitment to ethical data use.

\noindent \textbf{Participant Recruitment.} 
We recruited graduate students as annotators and required all participants to achieve an IELTS score of 6 or above. 
To mitigate potential biases stemming from participants’ geographical backgrounds, we minimized national differences in the dataset by focusing on human commonsense. 
All annotators provided informed consent and were compensated above the local minimum wage—\$10 per hour for standard annotators and \$20 per hour for senior annotators.

\noindent \textbf{Potential Risk.} After careful examination, we confirmed that our dataset does not contain any personal data (e.g., names, contacting information), and our data collection procedures adhere to ethical guidelines.

\section*{Limitation}

Due to the limited number of available domains online, we did not construct a large-scale training set. 
Future work should focus on expanding the dataset by incorporating additional data sources, such as synthesized data~\cite{hu2024agentgen}, to cover a broader range of domains.
Furthermore, although we conducted regular inspections to minimize the introduction of subjectivity into the dataset, the unavoidable influence of human subjectivity during manual annotation may introduce potential biases.


\nocite{langley00}

\bibliography{custom}

\newpage
\appendix
\onecolumn

\section{Benchmark Construction}
\subsection{Example}
\label{app:dataset-example}

\subsubsection{Domain Example}

\begin{lstlisting}[style=pddl,caption={Grid PDDL},label=lst:grid-domain]
(define (domain grid)
  (:requirements :strips)
  (:predicates (conn ?x ?y) (key-shape ?k ?s) (lock-shape ?x ?s)
	       (at ?r ?x ) (at-robot ?x) (place ?p) (key ?k) (shape ?s)
	       (locked ?x) (holding ?k)  (open ?x)  (arm-empty ))

  (:action unlock
    :parameters (?curpos ?lockpos ?key ?shape)
    :precondition (and (place ?curpos) (place ?lockpos) (key ?key)
		       (shape ?shape) (conn ?curpos ?lockpos)
		       (key-shape ?key ?shape) (lock-shape ?lockpos ?shape)
		       (at-robot ?curpos) (locked ?lockpos) (holding ?key))
    :effect (and (open ?lockpos) (not (locked ?lockpos))))

  (:action move
    :parameters (?curpos ?nextpos)
    :precondition (and (place ?curpos) (place ?nextpos) (at-robot ?curpos)
		       (conn ?curpos ?nextpos) (open ?nextpos))
    :effect (and (at-robot ?nextpos) (not (at-robot ?curpos))))

  (:action pickup
    :parameters (?curpos ?key)
    :precondition (and (place ?curpos) (key ?key) (at-robot ?curpos)
		       (at ?key ?curpos) (arm-empty ))
    :effect (and (holding ?key) (not (at ?key ?curpos)) (not (arm-empty ))))

  (:action pickup-and-loose
    :parameters (?curpos ?newkey ?oldkey)
    :precondition (and (place ?curpos) (key ?newkey) (key ?oldkey)
		       (at-robot ?curpos) (holding ?oldkey)
		       (at ?newkey ?curpos))
    :effect (and (holding ?newkey) (at ?oldkey ?curpos)
		 (not (holding ?oldkey)) (not (at ?newkey ?curpos))))

  (:action putdown
    :parameters (?curpos ?key)
    :precondition (and (place ?curpos) (key ?key) (at-robot ?curpos)
		       (holding ?key))
    :effect (and (arm-empty ) (at ?key ?curpos) (not (holding ?key))))
  )
\end{lstlisting}

\subsubsection{Abstract Description}
\label{app:diffdesc}

\noindent \textbf{General.}
This domain models a robot navigating a grid environment with the objective of unlocking doors and moving through the grid. The robot can carry keys that match the shape of locks to unlock doors. The environment includes places, keys with specific shapes, and doors (locks) with corresponding shapes that need to be unlocked.

\noindent \textbf{Predicates.}
The following predicates are used in the domain:
\begin{itemize}
    \item \texttt{(conn ?x ?y)}: Indicates a connection between two places \texttt{?x} and \texttt{?y}, allowing movement between them.
    \item \texttt{(key-shape ?k ?s)}: Indicates that key \texttt{?k} has shape \texttt{?s}.
    \item \texttt{(lock-shape ?x ?s)}: Indicates that lock (or door) at place \texttt{?x} has shape \texttt{?s}.
    \item \texttt{(at ?r ?x)}: Indicates that key \texttt{?r} is at place \texttt{?x}.
    \item \texttt{(at-robot ?x)}: Indicates that the robot is at place \texttt{?x}.
    \item \texttt{(place ?p)}: Indicates that \texttt{?p} is a place in the grid.
    \item \texttt{(key ?k)}: Indicates that \texttt{?k} is a key.
    \item \texttt{(shape ?s)}: Indicates that \texttt{?s} is a shape.
    \item \texttt{(locked ?x)}: Indicates that the place \texttt{?x} is locked.
    \item \texttt{(holding ?k)}: Indicates that the robot is holding key \texttt{?k}.
    \item \texttt{(open ?x)}: Indicates that the place \texttt{?x} is open.
    \item \texttt{(arm-empty)}: Indicates that the robot's arm is empty.
\end{itemize}

\noindent \textbf{Actions.}
The following actions are available in the domain:
\begin{itemize} 
    \item \texttt{unlock <?curpos> <?lockpos> <?key> <?shape>}: Allows the robot to unlock a door at place \texttt{<?lockpos>} using a key of a specific shape.
    \item \texttt{move <?curpos> <?nextpos>}: Allows the robot to move from place \texttt{<?curpos>} to place \texttt{<?nextpos>}.
    \item \texttt{pickup <?curpos> <?key>}: Allows the robot to pick up a key at its current location.
    \item \texttt{pickup-and-loose <?curpos> <?newkey> <?oldkey>}: Allows the robot to pick up a new key while dropping the one it was holding.
    \item \texttt{putdown <?curpos> <?key>}: Allows the robot to put down a key it is holding.
\end{itemize}

\subsubsection{Concrete Description}
\label{app:concrete_desc}

\noindent \textbf{General.}
This domain models a robot navigating a grid environment with the objective of unlocking doors and moving through the grid. The robot can carry keys that match the shape of locks to unlock doors. The environment includes places, keys with specific shapes, and doors (locks) with corresponding shapes that need to be unlocked.

\noindent \textbf{Predicates.}
The following predicates are used in the domain:
\begin{itemize}
    \item \texttt{(conn ?x ?y)}: Indicates a connection between two places \texttt{?x} and \texttt{?y}, allowing movement between them.
    \item \texttt{(key-shape ?k ?s)}: Indicates that key \texttt{?k} has shape \texttt{?s}.
    \item \texttt{(lock-shape ?x ?s)}: Indicates that lock (or door) at place \texttt{?x} has shape \texttt{?s}.
    \item \texttt{(at ?r ?x)}: Indicates that key \texttt{?r} is at place \texttt{?x}.
    \item \texttt{(at-robot ?x)}: Indicates that the robot is at place \texttt{?x}.
    \item \texttt{(place ?p)}: Indicates that \texttt{?p} is a place in the grid.
    \item \texttt{(key ?k)}: Indicates that \texttt{?k} is a key.
    \item \texttt{(shape ?s)}: Indicates that \texttt{?s} is a shape.
    \item \texttt{(locked ?x)}: Indicates that the place \texttt{?x} is locked.
    \item \texttt{(holding ?k)}: Indicates that the robot is holding key \texttt{?k}.
    \item \texttt{(open ?x)}: Indicates that the place \texttt{?x} is open.
    \item \texttt{(arm-empty)}: Indicates that the robot's arm is empty.
\end{itemize}

\noindent \textbf{Actions.} The following actions are available in the domain:
\begin{itemize}
\item \texttt{unlock <?curpos> <?lockpos> <?key> <?shape>}: Allows the robot to unlock a door at place \texttt{<?lockpos>} using a key of a specific shape if the robot is at place \texttt{<?curpos>}, the key matches the lock's shape, the robot is holding the key, there is a connection between \texttt{<?curpos>} and \texttt{<?lockpos>}, and the destination is locked. After the action, the lock is no longer locked.
\item \texttt{move <?curpos> <?nextpos>}: Allows the robot to move from place \texttt{<?curpos>} to place \texttt{<?nextpos>} if there is a connection between them and the destination is open. After the move, the robot is no longer at the original place.
\item \texttt{pickup <?curpos> <?key>}: Allows the robot to pick up a key at its current location if the robot's arm is empty and it is at the same place as the new key. After the action, the robot is holding the key, and the key is no longer at that location.
\item \texttt{pickup-and-loose <?curpos> <?newkey> <?oldkey>}: Allows the robot to pick up a new key while dropping the one it was holding if it is at the same place as the new key. After the action, the robot is holding the new key, and the old key is at the robot's current location.
\item \texttt{putdown <?curpos> <?key>}: Allows the robot to put down a key it is holding if it is at a specific place. After the action, the robot's arm is empty, and the key is at that location.
\end{itemize}

\subsection{Preliminary Experiment}
\label{app:self_eval}

The experimental results show that LLM's effectiveness in detecting PDDL semantic errors is limited, with an accuracy of 55.0\%, a precision of 56.2\%, a recall rate of 45.0\%, an F1 score of 50.0\%, and a ROC AUC of 55.0. ROC AUC indicates that the model is close to random performance, making it difficult to reliably distinguish between correct and incorrect PDDL domains. 
Below is the prompt used for LLMs to detect semantic errors in generated PDDL domains:\\
\lstset{
    basicstyle=\small\ttfamily, 
    breaklines=true, 
}
\begin{lstlisting}
You are an expert in automated planning systems and PDDL semantics. Your task is to evaluate whether the LLM are physically accurate models of the world or whether they don't make sense by detecting semantic errors in generated PDDL domain.
You need carefully analyze the following PDDL domain by comparing it to the pddl domain description, evaluate whether the generated pddl domain contains SEMANTIC ERRORS in these key aspects:
1. Predicates consistency.
2. Action parameters validity.
3. Action preconditions completeness.
4. Action effects logical consistency.
5. Consistency with the description.

An example of semantic error would be:
1. Missing precondition constraints (e.g. executing "unlock-door" without holding a key).
2. Contradictory effects (e.g. both adding and deleting the same predicate).
3. Incorrect predicate arguments (e.g. reversed parameter order).

Output Format:
{
"evaluation": "yes/no",
"error_type": "[MissingPrecond|IncorrectEffect|MissingPredicate|...]",
"confidence": "high/medium/low",
"evidence": "<specific code segment>",
"justification": "<short justification>"
}

PDDL Description:
{PDDL_DESCRIPTION}

Generated PDDL:
{PDDL_DOMAIN}
\end{lstlisting}

\subsection{More Details on Data Analysis}
\label{app:detailed_data_analysis}

\begin{figure}[htbp]
    \centering
    \includegraphics[width=\linewidth]{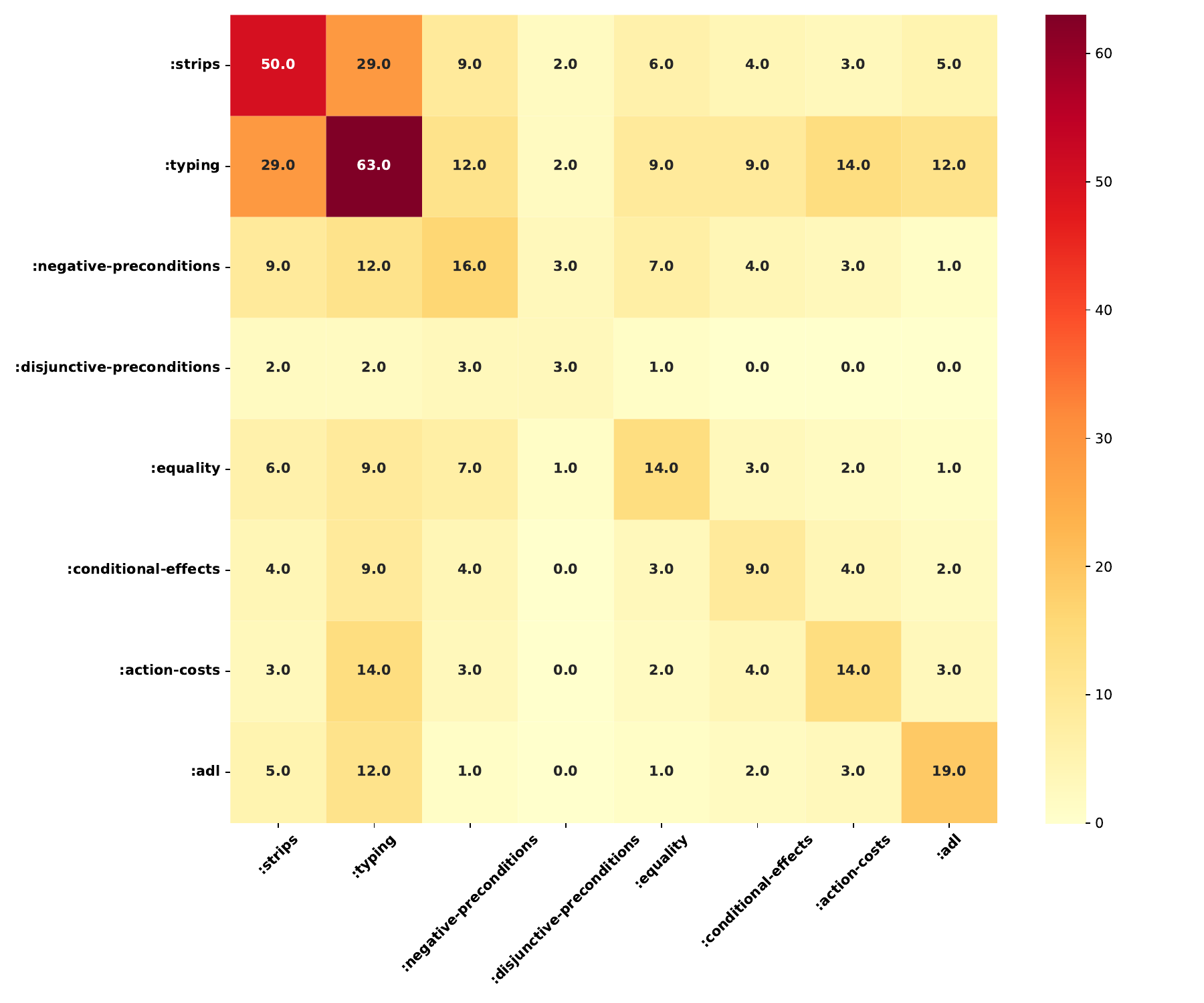}
    \caption{
    The co-occurrence matrix of requirements of \benchmark.
    }
    \label{fig:heatmap}
\end{figure}

Figure~\ref{fig:heatmap} shows the co-occurrence of PDDL requirements across domains, highlighting that \texttt{:typing} and \texttt{:strips} are the most prevalent features.

\section{More Details on Experiments}
\label{app:exp}

\subsection{Evaluation Metrics}
\label{app:eval_metric}

\noindent \textbf{Levenshtein Ratio.} 
The Levenshtein Ratio is a value between 0 and 1 that quantifies the similarity between two strings, such as a predicted PDDL domain and a golden PDDL domain. It is derived from the Levenshtein distance, which calculates the minimum number of character-level operations—insertions, deletions, or substitutions—needed to convert one string into the other. The ratio is then computed by dividing the Levenshtein distance by the length of the longer string, providing a measure of how closely the two strings match, where a value closer to 1 indicates high similarity and a value closer to 0 indicates significant differences.

\noindent \textbf{Component-wise F1 Scores.} 
The F1 score is mainly used to measure the similarity between the predicted PDDL domain and the golden PDDL domain, specifically including predicate F1 and action F1. The range of this score is from 0 to 1, which is the harmonic mean of precision and recall.


\subsection{Prompt Examples}
\label{app:prompt}

\subsubsection{Error Correction}
\label{app:error}

\begin{lstlisting}
I would like you to serve as an expert in PDDL, assisting me in correcting erroneous PDDL code. I will provide you with the incorrect PDDL along with the error messages returned by the system. You should output your thought process firstly. You MUST enclose the COMPLETE corrected PDDL within ```pddl```.
Here are some hints to help you debug the pddl domain file:
1. You should start by checking if all the essential domain constructs or domain definition constructs are present. Commonly included domains comprise:
    a. :domain declaration to name the domain.
    b. :requirements to specify the PDDL features used in the domain.
    c. :types to define any object types for categorizing entities in the planning problem.
    d. :constants (if necessary) to declare any objects that remain static throughout the planning problems.
    e. :predicates to define the properties and relations between objects that can change over time.
    f. :functions (if necessary) to define numeric functions for more complex domains.
    g. :action definitions for each action that agents can perform, including parameters, preconditions, and effects.
2. You need to check the number of parameters of each actions.
3. Having :typing in the domain indicates that it uses a hierarchy to organize objects. Therefore, it's crucial to clearly list all object types related to the planning task in a :types section.
4. '-' should not appear in :types.

Round 0
Incorrect PDDL:
(:action clean-up
    :parameters (?robot - robot ?robotTile - tile ?tileToBeCleaned - tile)
    :precondition (and 
        (robot-at ?robot ?robotTile)
        (up ?tileToBeCleaned ?robotTile)
        (clear ?tileToBeCleaned)
        (not (cleaned ?tileToBeCleaned))
    )
    :effect (and 
        (cleaned ?tileToBeCleaned)
    )
)

(:action clean-down
    :parameters (?robot - robot ?robotTile - tile ?tileToBeCleaned - tile)
    :precondition (and 
        (robot-at ?robot ?robotTile)
        (down ?tileToBeCleaned ?robotTile)
        (clear ?tileToBeCleaned)
        (not (cleaned ?tileToBeCleaned))
    )
    :effect (and 
        (cleaned ?tileToBeCleaned)
    )
)

(:action up
    :parameters (?robot - robot ?robotTile - tile ?moveToNextTile - tile)
    :precondition (and 
        (robot-at ?robot ?robotTile)
        (up ?moveToNextTile ?robotTile)
        (clear ?moveToNextTile)
    )
    :effect (and 
        (not (robot-at ?robot ?robotTile))
        (robot-at ?robot ?moveToNextTile)
    )
)

(:action down
    :parameters (?robot - robot ?robotTile - tile ?moveToNextTile - tile)
    :precondition (and 
        (robot-at ?robot ?robotTile)
        (down ?moveToNextTile ?robotTile)
        (clear ?moveToNextTile)
    )
    :effect (and 
        (not (robot-at ?robot ?robotTile))
        (robot-at ?robot ?moveToNextTile)
    )
)

(:action right
    :parameters (?robot - robot ?robotTile - tile ?moveToNextTile - tile)
    :precondition (and 
        (robot-at ?robot ?robotTile)
        (right ?moveToNextTile ?robotTile)
        (clear ?moveToNextTile)
    )
    :effect (and 
        (not (robot-at ?robot ?robotTile))
        (robot-at ?robot ?moveToNextTile)
    )
)

(:action left
    :parameters (?robot - robot ?robotTile - tile ?moveToNextTile - tile)
    :precondition (and 
        (robot-at ?robot ?robotTile)
        (left ?moveToNextTile ?robotTile)
        (clear ?moveToNextTile)
    )
    :effect (and 
        (not (robot-at ?robot ?robotTile))
        (robot-at ?robot ?moveToNextTile)
    )
)
Error Information:
ParsingError: line 1:1 mismatched input ':action' expecting 'define'
Corrected PDDL:
\end{lstlisting}

\subsubsection{Zero-Shot Prompt}

\begin{lstlisting}
You are tasked with converting a given Planning Domain Definition Language (PDDL) domain description into its corresponding formal PDDL domain. The description will outline the essential components of the domains. 
Your output should be a well-structured PDDL domain that accurately represents the given description, adhering to the syntax and semantics of PDDL.
Your output pddl domain must be enclosed in ```pddl```.

You need to generate the corresponding domain pddl for the following description.
    
PDDL Domain Description:
### General
This domain is designed for a robot tasked with cleaning floor tiles. The robot can move in four directions (up, down, right, left) relative to its current position on a grid of tiles. The goal is to clean all the specified tiles by moving to them and performing a cleaning action.

### Types
- **robot**: Represents the robot that performs the cleaning.
- **tile**: Represents the individual tiles on the floor that may need to be cleaned.

### Predicates
- **(robot-at ?robot - robot ?robotTile - tile)**: Indicates that the robot is currently at a specific tile.
- **(up ?tileAbove - tile ?tileBelow - tile)**: Indicates that one tile is directly above another.
- **(down ?tileBelow - tile ?tileAbove - tile)**: Indicates that one tile is directly below another.
- **(right ?tileOnRight - tile ?tileOnLeft - tile)**: Indicates that one tile is directly to the right of another.
- **(left ?tileOnLeft - tile ?tileOnRight - tile)**: Indicates that one tile is directly to the left of another.
- **(clear ?clearedTile - tile)**: Indicates that a tile is clear and robot can move there.
- **(cleaned ?cleanedTile - tile)**: Indicates that a tile has been cleaned.

### Actions
- **clean-up <?robot> <?robotTile> <?tileToBeCleaned>**: Allows the robot (?robot) to clean a tile (?tileToBeCleaned) that is directly above its current position (?robotTile).  

- **clean-down <?robot> <?robotTile> <?tileToBeCleaned>**: Allows the robot (?robot) to clean a tile (?tileToBeCleaned) that is directly below its current position (?robotTile).  

- **up <?robot> <?robotTile> <?moveToNextTile>**: Moves the robot (?robot) to a tile (?moveToNextTile) directly above its current position (?robotTile).  

- **down <?robot> <?robotTile> <?moveToNextTile>**: Moves the robot (?robot) to a tile (?moveToNextTile) directly below its current position (?robotTile).  

- **right <?robot> <?robotTile> <?moveToNextTile>**: Moves the robot (?robot) to a tile (?moveToNextTile) directly to the right of its current position (?robotTile).  

- **left <?robot> <?robotTile> <?moveToNextTile>**: Moves the robot (?robot) to a tile (?moveToNextTile) directly to the left of its current position (?robotTile).
PDDL Domain:
Let's think step by step.
\end{lstlisting}

\subsubsection{Few-Shot Prompt}

\begin{lstlisting}
You are tasked with converting a given Planning Domain Definition Language (PDDL) domain description into its corresponding formal PDDL domain. The description will outline the essential components of the domains. Your output should be a well-structured PDDL domain that accurately represents the given description, adhering to the syntax and semantics of PDDL.
Your output must strictly adhere to the format exemplified below. 
Here are some examples:

Example 0:
## PDDL Domain Description
### General
You are a robot equipped with a gripper mechanism, designed to move and manipulate balls between different rooms. The domain focuses on the robot's ability to navigate rooms, pick up balls, and drop them in designated locations.
### Types
- **room**: Represents the different rooms within the environment.
- **ball**: Represents the objects that the robot can pick up and move.
- **gripper**: Represents the robot's mechanism for holding balls.
### Predicates
- **(at-robby ?r - room)**: Indicates that Robby, the robot, is currently in room ?r.
- **(at ?b - ball ?r - room)**: Indicates that ball ?b is located in room ?r.
- **(free ?g - gripper)**: Indicates that the gripper ?g is not currently holding any ball.
- **(carry ?o - ball ?g - gripper)**: Indicates that the gripper ?g is carrying ball ?o.
### Actions
- **move <?from> <?to>**: Allows Robby to move from one room to another.  
- **pick <?obj> <?room> <?gripper>**: Enables Robby to pick up a ball in a room using its gripper.  
- **drop <?obj> <?room> <?gripper>**: Allows Robby to drop a ball it is carrying into a room.

## PDDL Domain
```pddl
(define (domain gripper-strips)
	(:types 
		room - object
		ball - object
		gripper - object
		)
   (:predicates
		(at-robby ?r - room)
		(at ?b - ball ?r - room)
		(free ?g - gripper)
		(carry ?o - ball ?g - gripper))
   (:action move
       :parameters  (?from - room ?to - room)
       :precondition (and (at-robby ?from))
       :effect (and  (at-robby ?to)
		     (not (at-robby ?from))))
   (:action pick
       :parameters (?obj - ball ?room - room ?gripper - gripper)
       :precondition  (and 
			    (at ?obj ?room) (at-robby ?room) (free ?gripper))
       :effect (and (carry ?obj ?gripper)
		    (not (at ?obj ?room)) 
		    (not (free ?gripper))))
   (:action drop
       :parameters  (?obj - ball ?room - room ?gripper - gripper)
       :precondition  (and 
			    (carry ?obj ?gripper) (at-robby ?room))
       :effect (and (at ?obj ?room)
		    (free ?gripper)
		    (not (carry ?obj ?gripper)))))
```

Example 1:
## PDDL Domain Description
### General
This domain represents a simplified version of the classic "blocks world" problem, where a robot arm can stack and unstack blocks. The domain includes actions for picking up blocks from the table, putting down blocks onto the table, stacking blocks on top of each other, and unstacking them. The goal is to manipulate the blocks to achieve a specified configuration.
### Predicates
(clear ?x): Indicates that there is no block on top of block ?x, making it accessible for stacking or picking up.
(on-table ?x): Indicates that block ?x is directly on the table.
(arm-empty): Indicates that the robot's arm is not holding any block.
(holding ?x): Indicates that the robot's arm is currently holding block ?x.
(on ?x ?y): Indicates that block ?x is directly on top of block ?y.
### Actions
- **pickup <?ob>**: Picks up an object (?ob) from the table.  
- **putdown <?ob>**: Puts down an object (?ob) onto the table.  
- **stack <?ob> <?underob>**: Stacks an object (?ob) on top of another object (?underob), making the robot arm empty.  
- **unstack <?ob> <?underob>**: Unstacks an object (?ob) from another object (?underob), making the robot arm no longer empty.

## PDDL Domain:
```pddl
(define (domain blocksworld)
  (:requirements :strips)
(:predicates (clear ?x)
             (on-table ?x)
             (arm-empty)
             (holding ?x)
             (on ?x ?y))

(:action pickup
  :parameters (?ob)
  :precondition (and (clear ?ob) (on-table ?ob) (arm-empty))
  :effect (and (holding ?ob) (not (clear ?ob)) (not (on-table ?ob)) 
               (not (arm-empty))))

(:action putdown
  :parameters  (?ob)
  :precondition (holding ?ob)
  :effect (and (clear ?ob) (arm-empty) (on-table ?ob) 
               (not (holding ?ob))))

(:action stack
  :parameters  (?ob ?underob)
  :precondition (and (clear ?underob) (holding ?ob))
  :effect (and (arm-empty) (clear ?ob) (on ?ob ?underob)
               (not (clear ?underob)) (not (holding ?ob))))

(:action unstack
  :parameters  (?ob ?underob)
  :precondition (and (on ?ob ?underob) (clear ?ob) (arm-empty))
  :effect (and (holding ?ob) (clear ?underob)
               (not (on ?ob ?underob)) (not (clear ?ob)) (not (arm-empty)))))
```

You need to generate the corresponding domain pddl for the following description.
    
## PDDL Domain Description
### General
This domain is designed for a robot tasked with cleaning floor tiles. The robot can move in four directions (up, down, right, left) relative to its current position on a grid of tiles. The goal is to clean all the specified tiles by moving to them and performing a cleaning action.

### Types
- **robot**: Represents the robot that performs the cleaning.
- **tile**: Represents the individual tiles on the floor that may need to be cleaned.

### Predicates
- **(robot-at ?robot - robot ?robotTile - tile)**: Indicates that the robot is currently at a specific tile.
- **(up ?tileAbove - tile ?tileBelow - tile)**: Indicates that one tile is directly above another.
- **(down ?tileBelow - tile ?tileAbove - tile)**: Indicates that one tile is directly below another.
- **(right ?tileOnRight - tile ?tileOnLeft - tile)**: Indicates that one tile is directly to the right of another.
- **(left ?tileOnLeft - tile ?tileOnRight - tile)**: Indicates that one tile is directly to the left of another.
- **(clear ?clearedTile - tile)**: Indicates that a tile is clear and robot can move there.
- **(cleaned ?cleanedTile - tile)**: Indicates that a tile has been cleaned.

### Actions
- **clean-up <?robot> <?robotTile> <?tileToBeCleaned>**: Allows the robot (?robot) to clean a tile (?tileToBeCleaned) that is directly above its current position (?robotTile).  

- **clean-down <?robot> <?robotTile> <?tileToBeCleaned>**: Allows the robot (?robot) to clean a tile (?tileToBeCleaned) that is directly below its current position (?robotTile).  

- **up <?robot> <?robotTile> <?moveToNextTile>**: Moves the robot (?robot) to a tile (?moveToNextTile) directly above its current position (?robotTile).  

- **down <?robot> <?robotTile> <?moveToNextTile>**: Moves the robot (?robot) to a tile (?moveToNextTile) directly below its current position (?robotTile).  

- **right <?robot> <?robotTile> <?moveToNextTile>**: Moves the robot (?robot) to a tile (?moveToNextTile) directly to the right of its current position (?robotTile).  

- **left <?robot> <?robotTile> <?moveToNextTile>**: Moves the robot (?robot) to a tile (?moveToNextTile) directly to the left of its current position (?robotTile).
## PDDL Domain
\end{lstlisting}

\section{More Details on Analysis}
\label{app:analysis_detail}

\subsection{Overall}


\begin{table}[htbp]
\centering
\caption{Distribution of error types of \texttt{claude-3.5-sonnect} on \benchmark under few-shot setting.}
\label{tab:error_distribution}
\begin{tabular}{l|c|c}
\toprule
 & \textbf{Proportion (\%)} & \textbf{Number} \\ \midrule
\textbf{Correct} & 23.76 & 24 \\ 
\textbf{Syntax Error} & 11.88 & 12 \\ 
\textbf{Semantic Error} & 64.36 & 65 \\ \midrule
\textbf{All} & 100.00 & 101 \\ 
\bottomrule
\end{tabular}
\end{table}

The overall distribution for syntax errors and semantic errors is presented in Table~\ref{tab:error_distribution}.

\subsection{Syntax Error}

\begin{table*}[htbp]
\centering
\caption{Distribution of Syntax Errors in PDDL Generation (Total Samples: 66, a task may have 1 to 4 samples.)}
\label{tab:syntax_errors}
\resizebox{\textwidth}{!}{
\begin{tabular}{l|l|c}
\toprule
\textbf{Syntax Error} & \textbf{Explanation} & \textbf{Proportion (\%)} \\ 
\midrule
UndefinedDomainName & Missing mandatory \texttt{(define (domain ...))} declaration in PDDL header & 33.33 \\ 
IncorrectParentheses & Invalid empty/mismatched parentheses & 3.03 \\ 
UndefinedConstant & Reference to undeclared constants in predicates or actions & 13.64 \\ 
MissingRequirements & Absence of required PDDL extension declarations (e.g., \texttt{:action-costs}) & 22.73 \\ 
UndefinedType & Undeclared parent type in hierarchical type definitions & 18.18 \\ 
UnsupportedFeature & Use of parser-incompatible language features (e.g., \texttt{either} types) & 3.03 \\ 
TypeMismatch & Parameter type conflict with declared type constraints & 1.52 \\ 
UndefinedVariable & Undeclared variables in action preconditions/effects & 1.52 \\ 
DuplicateDefinition & Multiple declarations of identical domain elements & 3.03 \\ 
\bottomrule
\end{tabular}
}
\end{table*}

The distribution and detailed explanation of each syntax error type are presented in Table~\ref{tab:syntax_errors}.

\subsection{Semantic Error}
\begin{table*}[htbp]
\centering
\caption{Distribution of Semantic Errors in PDDL Generation (Total Samples: 91, a task may have multiple semantic errors.)}
\label{tab:semantic_errors}
\resizebox{\textwidth}{!}{
\begin{tabular}{llc}
\toprule
\textbf{Semantic Error} & \textbf{Explanation} & \textbf{Proportion (\%)} \\ 
\midrule
\textbf{DisobeyDescription} & Direct violation of semantic requirements explicitly stated in the task description. & \textbf{14.29} \\ 
\quad IncorrectPredicate & Incorrect or missing the declaration of predicates. & 6.59 \\ 
\quad IncorrectAction & Incorrect or missing the declaration of actions. & 7.69 \\ 
\midrule
\textbf{IncompleteModeling} & Incomplete world modeling compared to basic requirements. & \textbf{58.24} \\ 
\quad IncorrectPrecondition & The precondition of the action is deficient or incorrect. & 29.67 \\ 
\quad IncorrectEffect & The effect of the action is deficient or incorrect. & 28.57 \\ 
\midrule
\textbf{RedundantSpecifications} & Predicted domain includes superfluous preconditions or effects. & \textbf{17.58} \\ 
\quad RedundantPrecondition & Predicted domain includes superfluous preconditions. & 10.99 \\ 
\quad RedundantEffect & Predicted domain includes superfluous effects. & 6.59 \\ 
\midrule
\textbf{SurfaceDivergence} & Surface variations preserving semantic equivalence with ground truth. & \textbf{9.89} \\ 
\bottomrule
\end{tabular}
}
\end{table*}

The distribution and detailed explanation of each semantic error type are presented in Table~\ref{tab:semantic_errors}.

\section{More Experimental Results}

\subsection{Experimental Results with Concrete Description}
\label{app:concrete_results}

\begin{table*}[htbp]
\centering
\caption{Performance comparison of different LLMs on \benchmark using concrete domain description. EC\textsubscript{k} denotes the setting where models are allowed k correction attempts (EC\textsubscript{0}: zero-shot without correction, EC\textsubscript{3}: with 3 correction attempts).}
\label{tab:main-concrete}
\resizebox{\textwidth}{!}{%
\begin{tabular}{l|c|cc|cc|cc|cc|cc|cc}
\toprule
\multirow{2}{*}{\textbf{Model Family}} & \multirow{2}{*}{\textbf{Version}} & \multicolumn{2}{c|}{\textbf{\exec} $\uparrow$} & \multicolumn{2}{c|}{\textbf{\simmetric} $\uparrow$} & \multicolumn{2}{c|}{\textbf{\fonepred} $\uparrow$} & \multicolumn{2}{c|}{\textbf{\foneparam} $\uparrow$} & \multicolumn{2}{c|}{\textbf{\foneprecond} $\uparrow$} & \multicolumn{2}{c}{\textbf{\foneeff} $\uparrow$} \\
\cmidrule{3-14} & & \textbf{EC\textsubscript{0}} & \textbf{EC\textsubscript{3}} & \textbf{EC\textsubscript{0}} & \textbf{EC\textsubscript{3}} & \textbf{EC\textsubscript{0}} & \textbf{EC\textsubscript{3}} & \textbf{EC\textsubscript{0}} & \textbf{EC\textsubscript{3}} & \textbf{EC\textsubscript{0}} & \textbf{EC\textsubscript{3}} & \textbf{EC\textsubscript{0}} & \textbf{EC\textsubscript{3}} \\
\midrule
\multirow{1}{*}{\textsc{GPT-4}} & {\small\texttt{gpt-4o}} & 60.4 & 75.2 & 90.7 & 90.3 & 59.4 & 71.8 & 57.1 & 69.1 & 55.3 & 65.1 & 54.1 & 65.2 \\
\midrule
\multirow{1}{*}{\textsc{GPT-3.5}} & {\small\texttt{turbo-0125}} & 53.5 & 68.3 & 89.0 & 88.7 & 52.9 & 66.7 & 50.3 & 64.6 & 45.1 & 58.0 & 46.5 & 59.9 \\
\midrule
\textsc{Claude-3.5} & {\small\texttt{sonnet}} & 64.4 & 84.2 & 84.7 & 77.6 & 64.4 & 80.7 & 55.0 & 67.5 & 53.3 & 65.0 & 53.3 & 64.8\\
\midrule
\multirow{2}{*}{\textsc{LLaMA-2}} & {\small\texttt{7b-instruct}} & 0.0 & 0.0 & 48.4 & 32.4 & 0.0 & 0.0 & 0.0 & 0.0 & 0.0 & 0.0 & 0.0 & 0.0 \\
 & {\small\texttt{70b-instruct}} & 0.0 & 0.0 & 53.5 & 52.5 & 0.0 & 0.0 & 0.0 & 0.0 & 0.0 & 0.0 & 0.0 & 0.0 \\
\midrule
\multirow{2}{*}{\textsc{LLaMA-3.1}} & {\small\texttt{8b-instruct}} & 0.0 & 1.0 & 84.1 & 83.2 & 0.0 & 0.0 & 0.0 & 0.0 & 0.0 & 0.0 & 0.0 & 0.0 \\
 & {\small\texttt{70b-instruct}} & 1.0 & 1.0 & 89.7 & 85.4 & 1.0 & 1.0 & 1.0 & 1.0 & 1.0 & 1.0 & 1.0 & 1.0 \\
\midrule
\multirow{1}{*}{\textsc{DeepSeek}} & {\small\texttt{deepseek-v3}} & 58.4 & 80.2 & 90.1 & 89.3 & 58.1 & 76.4 & 56.2 & 73.5 & 53.4 & 66.0 & 53.5 & 67.6 \\ 
\bottomrule
\end{tabular}%
}
\end{table*}


\end{document}

%% file: section/experiment.tex
\input{table/main}